%% file: main.tex
\documentclass[10pt,letterpaper]{article}
\usepackage[letterpaper,textwidth=5.5in,textheight=9in,top=1in,hcentering,footskip=30pt]{geometry}
\usepackage[T1]{fontenc}
\usepackage{amsmath}
\usepackage{mathtools}
\usepackage{amsthm}
\usepackage{newtxtext}
\usepackage{newtxmath}
\usepackage{bm,nicefrac,mathrsfs}
\usepackage{microtype}
\usepackage{verbatim}
\usepackage{setspace}
\usepackage{algorithm,algorithmic,refcount}
\usepackage{booktabs}
\usepackage{placeins}
\usepackage{graphicx}
\usepackage[dvipsnames]{xcolor}
\usepackage{xargs}
\usepackage[backend=biber,style=numeric,sorting=nyt,sortcites=true,maxbibnames=99,giveninits=true,doi=true,url=false,isbn=false,eprint=false]{biblatex}
\makeatletter
\renewcommand{\section}{\@startsection{section}{1}{\z@}%
  {-2.0ex plus -0.5ex minus -.2ex}{1.5ex plus 0.3ex minus .2ex}{\large\bfseries\raggedright}}
\renewcommand{\subsection}{\@startsection{subsection}{2}{\z@}%
  {-1.8ex plus -0.5ex minus -.2ex}{0.8ex plus .2ex}{\normalsize\bfseries\raggedright}}
\renewcommand{\subsubsection}{\@startsection{subsubsection}{3}{\z@}%
  {-1.5ex plus -0.5ex minus -.2ex}{0.5ex plus .2ex}{\normalsize\bfseries\raggedright}}
\newcommand{\toptitlebar}{\hrule height 4pt\vskip .25in\vskip-\parskip}
\newcommand{\bottomtitlebar}{\vskip .29in\vskip-\parskip\hrule height 1pt\vskip .09in}
\newcommand{\@authorsep}{\end{tabular}\hfil\linebreak[0]\hfil
  \begin{tabular}[t]{c}\bfseries\rule{\z@}{24pt}\ignorespaces}
\renewcommand{\maketitle}{\par\begingroup
  \renewcommand{\thefootnote}{\fnsymbol{footnote}}%
  \def\@makefnmark{\hbox to\z@{$^{\@thefnmark}$\hss}}%
  \long\def\@makefntext##1{\parindent 1em\noindent\hbox to 1.8em{\hss$\m@th^{\@thefnmark}$}##1}%
  \@maketitle\@thanks
  \endgroup
  \setcounter{footnote}{0}%
  \global\let\thanks\relax\global\let\maketitle\relax}
\def\@maketitle{\vbox{\hsize\textwidth\linewidth\hsize\let\And\@authorsep
  \vskip 0.1in\toptitlebar\centering
  {\LARGE\bfseries\@title\par}\bottomtitlebar
  \begin{tabular}[t]{c}\bfseries\rule{\z@}{24pt}\@author\end{tabular}%
  \vskip 0.3in minus 0.1in}}
\renewenvironment{abstract}{\vskip .075in\centerline{\large\bfseries Abstract}\vspace{0.5ex}%
  \list{}{\leftmargin 0.5in\rightmargin\leftmargin}\item\relax}{\endlist\vskip 1ex}
\makeatother

\newtheorem{theorem}{Theorem}
\newtheorem{lemma}{Lemma}
\newtheorem{proposition}{Proposition}

\newtheorem{definition}{Definition}
\newtheorem{remark}{Remark}
\newtheorem{assumption}{Assumption}
\numberwithin{equation}{section}

\newcommand{\keywords}[1]{\par\vskip 1ex\noindent\textbf{Keywords:} #1}

\usepackage[most]{tcolorbox}
\definecolor{cgreen}{HTML}{0F7A58}
\definecolor{cpurple}{HTML}{534AB7}
\definecolor{cgrey}{HTML}{6B6B6B}
\definecolor{cbga}{HTML}{F4F8F6}
\definecolor{cbgb}{HTML}{F5F6FB}
\definecolor{cbgc}{HTML}{F6F6F6}
\tcbset{
  stepgreen/.style={colframe=cgreen,colback=cbga,coltitle=cgreen,colbacktitle=cbga},
  steppurple/.style={colframe=cpurple,colback=cbgb,coltitle=cpurple,colbacktitle=cbgb},
  stepgrey/.style={colframe=cgrey,colback=cbgc,coltitle=cgrey,colbacktitle=cbgc},
}
\newtcolorbox{stepbox}[2][stepgreen]{enhanced,boxrule=0.6pt,arc=2pt,left=6pt,right=6pt,
  top=2pt,bottom=4pt,toptitle=3pt,bottomtitle=0pt,fonttitle=\bfseries,
  before skip=6pt,after skip=10pt,title={#2},#1, titlerule=0pt}

\usepackage[hidelinks]{hyperref}

\newcommand{\zpop}{\hat z_{\mathsf{pop}}}

\newcommand{\Lyap}{\mathcal V}
\newcommand{\Dzeta}{D\zeta}
\newcommand{\nin}{L}
\newcommand{\Kmin}{\mathcal K_{\min}}
\newcommand{\Kmax}{\mathcal K_{\max}}
\newcommand{\Lip}[1]{L_{#1}}
\newcommand{\cdec}{c_{\mathsf{dec}}}
\newcommand{\cgrad}{c_{\mathsf{grad}}}
\newcommand{\cLyap}{c_{\mathcal V}}
\newcommand{\Sigtot}{\Sigma^{\mathsf{tot}}}
\newcommand{\Ntot}{N_{\mathrm{tot}}}
\newcommand{\Reals}{\mathbb R}
\newcommand{\Exp}{\mathbb E}
\newcommand{\tr}{\operatorname{tr}}
\newcommand{\dS}{n_{\mathcal S}}
\newcommand{\vq}{\mathsf q}
\newcommand{\Fmf}{\mathcal F}
\newcommand{\Flow}{\Phi^{\Fmf}}

\input{graphics/numbers.tex}

\input{graphics/story_numbers.tex}

\input{graphics/theory_numbers.tex}

\input{graphics/encoder_numbers.tex}

\title{From the ELBO to Maximum Likelihood:\\ Convergence of Variational Autoencoders\\ for Nonlinear Mixed-Effects Modeling}

\author{%
  Jan Rohleff\thanks{Corresponding author.}\\
  Department of Mathematics and Statistics\\
  University of Konstanz, Germany\\
  \href{mailto:jan.rohleff@uni-konstanz.de}{\texttt{jan.rohleff@uni-konstanz.de}}
  \And
  Johannes Schropp\\
  Department of Mathematics and Statistics\\
  University of Konstanz, Germany\\
  \href{mailto:johannes.schropp@uni-konstanz.de}{\texttt{johannes.schropp@uni-konstanz.de}}
}

\hypersetup{
  pdftitle={From the ELBO to Maximum Likelihood: Convergence of Variational Autoencoders for Nonlinear Mixed-Effects Modeling},
  pdfauthor={Jan Rohleff, Johannes Schropp}
}

\begin{document}

\maketitle

\begin{abstract}
  \noindent
Nonlinear mixed-effects (NLME) models are the central statistical models of pharmacometrics,
where they support dose selection, individualized dosing and regulatory decisions in drug
development. Statistically, NLME modeling is a marginalized likelihood problem. Variational
autoencoders (VAEs) solve such problems with a neural network encoder that amortizes the
approximation of the intractable posterior distribution, and allow the integration of AI components
into NLME modeling. This paper provides the convergence analysis of the VAE-NLME algorithm. First,
its version for the evidence lower bound (ELBO) is proven to converge almost surely to the set of stationary
points of the ELBO, without assuming that the encoder is trained exactly. Second, the final VAE-NLME
algorithm trains the encoder on the ELBO by gradient-based optimization and uses it as proposal for
importance-weighted sufficient statistics. It converges almost surely to the set of stationary points
of the marginal log-likelihood, without any assumption on the training of the encoder. Thus, it
combines the fast gradient-based training of VAEs with the limit of maximum likelihood estimation.

\keywords{Variational Autoencoder, Maximum Likelihood Estimation, Stochastic Approximation, Nonlinear Mixed-Effects Modeling, Pharmacometrics, AI}
\end{abstract}

\section{Introduction}
\label{Section:1}
Variational autoencoders (VAEs)
\cite{KingmaAutoEncodingVariationalBayes14,RezendeStochasticBackpropagationApproximate14} are a
central method of generative artificial intelligence (AI). They fit latent variable models with an
encoder network that amortizes posterior inference. This paper analyzes their convergence for
nonlinear mixed-effects (NLME) models
\cite{SheinerEstimationPopulationCharacteristics77,MouldBasicConceptsPopulation12}, the central
models of pharmacometrics (PMX) for regulatory decisions in drug development
\cite{BhattaramImpactPharmacometricsDrug05,MadabushiReviewRoleModelInformed22}.

\noindent
An NLME model is a hierarchical latent variable model. Estimating the population parameters
$\theta$ from the observations $x$ of all patients requires integrating out the individual parameters $z$, i.e., NLME modeling is a
marginalized likelihood problem,
\begin{align*}
  \max_{\theta}\ \log\int p(x|z;\theta)\,p(z;\theta)\,dz .
\end{align*}
In PMX, NLME models are mostly given by nonlinear
ordinary differential equations (ODEs), so that the integral has no closed form. The industry gold standards are first-order conditional estimation with interaction
(FOCEI), which linearizes the model
\cite{LindstromNonlinearMixedEffects90,WangDerivationVariousNONMEM07}, and the stochastic
approximation expectation maximization (SAEM) algorithm, which combines the expectation-maximization
(EM) algorithm \cite{DempsterMaximumLikelihoodIncomplete77} with Markov chain Monte Carlo (MCMC)
sampling \cite{DelyonConvergenceStochasticApproximation99,KuhnCouplingStochasticApproximation04}.

\noindent
To automate model development and to integrate more patient data, such as medical images (e.g., CT
scans), there is growing interest in incorporating AI models with neural network components into NLME modeling
\cite{RichardsonMachineLearningApproach25,KochPharmacometricsMachineLearning20,LuNeuralODEPharmacokineticsModeling21,BramLowdimensionalNeuralOrdinary25},
for which the established methods FOCEI and SAEM are not well suited
\cite{RohleffEvolutionNonlinearMixed26}. Therefore, VAEs were adapted to NLME models
\cite{RohleffRedefiningParameter25,RohleffVariationalAutoencodersNeural26}. Related approaches to adapt
VAEs to NLME models have also been explored in
\cite{ArrudaAmortizedApproachNonLinear24,LiVariationalAutoencoderInference26,TarekFittingLargeNonlinear26,BaazEmpiricalBayesVariational26}.
The resulting evidence lower bound (ELBO) \mbox{VAE-NLME} algorithm replaces the MCMC step of SAEM by a neural network encoder
trained by stochastic gradients. It is fast and is well-suited for AI applications, but it
has never been analyzed mathematically. Moreover, it maximizes only the ELBO,
so that its limit, if it exists, is a maximum of the ELBO instead of the maximum likelihood
estimator. In PMX, however, the likelihood is the hard currency for the quality of a model, e.g., for
model comparison \cite{DelattreNoteBICMixedeffects14}. An algorithm that solves the likelihood
problem only approximately is not usable in practice, however fast it is. Its regulatory use also
requires a mathematical validation \cite{FDAConsiderationsUseArtificial25,EMAReflectionPaperUse24}.
First, does the algorithm converge at all, although convergence cannot be shown in general for the
training of neural networks? Second, can it be extended to converge to the maximum likelihood
estimator while keeping its gradient-based training?

\noindent
The key idea is to use the ELBO and the likelihood for different purposes. The ELBO trains the
encoder by stochastic gradients, and the encoder serves as proposal for importance-weighted
sufficient statistics, which move the limit from the ELBO to the marginal log-likelihood. Unlike SAEM
\cite{DelyonConvergenceStochasticApproximation99,KuhnCouplingStochasticApproximation04}, this needs
no samples from the posterior distribution, and unlike biased stochastic approximation
in general \cite{KarimiNonasymptoticAnalysisBiased19}, it controls the bias without any assumption
on the training of the encoder. Hence, the algorithm converges to the maximum likelihood estimator while retaining the computational efficiency of gradient-based optimization \cite{RohleffBenchmarkingVariationalAutoencoder26}. Section~\ref{sec:convergence} answers
both questions by three theorems, whose proofs by stochastic approximation
\cite{StochasticApproximationRecursive03,BorkarStochasticApproximationDynamical23} are outlined in
Section~\ref{sec:proofs}. Section~\ref{sec:experiments} confirms the theorems numerically.

\noindent
\emph{Related work.} The update of the sufficient statistics is that of online EM
\cite{CappeOnlineExpectationMaximization09}, and SAEM with biased MCMC samplers is analyzed in
\cite{GruffazStochasticApproximationBiased24}. Black-box variational inference is proven to converge
for fixed model parameters without an encoder
\cite{DomkeProvableSmoothnessGuarantees20,DomkeProvableConvergenceGuarantees23,KimConvergenceBlackBoxVariational23},
and the loss of accuracy caused by an encoder, the amortization gap, is studied in
\cite{CremerInferenceSuboptimalityVariational18,MargossianAmortizedVariationalInference24}.
Importance sampling with the encoder as proposal is also used in reweighted wake-sleep
\cite{BornscheinReweightedWakeSleep15} and importance weighted autoencoders
\cite{BurdaImportanceWeightedAutoencoders16}. 

\section{Background}
\label{sec:background}
This section recalls the NLME model and the VAE framework for NLME modeling from
\cite{RohleffRedefiningParameter25}.
\subsection{NLME Models}
\label{Section:2}
Consider a dataset of $N$ individuals. For individual $i\in\{1,\dots,N\}$ and observation
$j\in\{1,\dots,n_i\}$, the observation $x_{ij}\in\Reals$ at the known variable $t_{ij}\in\Reals^p$,
e.g., the time, is modeled as
\begin{align}
  \label{eq:NLME_model}
    x_{ij} = f(t_{ij}, \phi_i) + g(t_{ij}, \phi_i)\,\sigma\,\varepsilon_{ij}, \quad 1 \leq i \leq N,\, 1 \leq j \leq n_i.
\end{align}
Here, $f$ is the nonlinear structural model, $g$ the residual error model and $\sigma>0$ the
residual error parameter.
\subsubsection{Individual Parameters}
The individual parameters $\phi_i\in\Phi$, where $\Phi\subseteq\Reals^{n_z}$ denotes the open set of
their admissible values, are given by the individual model (prior distribution)
\begin{align*}
  \begin{aligned}
  h(\phi_i) &= z_i, \\
    z_i &= z_\mathsf{pop} + \beta c_i + \eta_i,
    \qquad \eta_i \overset{\text{i.i.d.}}{\sim} \mathcal{N}(0,\Omega),
      \end{aligned}
\end{align*}
where $h:\Phi\to\Reals^{n_z}$ is a smooth invertible transformation to adjust the prior to non-Gaussian distributions,
$z_{\mathsf{pop}}\in\Reals^{n_z}$ the fixed effects, $c_i=(c_{i1},\dots,c_{in_c})^\top\in\Reals^{n_c}$
the covariates of individual $i$, $\beta=(\beta_1,\dots,\beta_{n_c})\in\Reals^{n_z\times n_c}$ the
matrix of covariate effects, and $\eta_i$ the random effects with positive definite covariance matrix
$\Omega\in\Reals^{n_z\times n_z}$, which describes the inter-individual variability.
\\
Collecting the fixed effects and the $n_\beta$ estimated covariate effects in
\begin{align*}
\hat z_{\mathsf{pop}} &= (z_{\mathsf{pop}},\beta)\in\Reals^{n_z+n_\beta},
\end{align*}
the individual model reads
\begin{align}
  \label{eq:individual_parameter_matrix}
z_i = C_i\hat z_{\mathsf{pop}}+\eta_i,
\quad
\eta_i &\overset{\mathrm{i.i.d.}}{\sim}\mathcal N(0,\Omega),
\end{align}
with a design matrix $C_i$ of full row rank $n_z$.
\subsubsection{Error Model}
The residual errors $\varepsilon_{ij}\sim\mathcal N(0,1)$ are independent, and independent of
$\eta_i$. Hence, $x_{ij}$ given $z_i$ is normally distributed with mean $f(t_{ij},\phi_i)$ and
standard deviation $g(t_{ij},\phi_i)\sigma$.
\subsubsection{Marginal Likelihood and Objective}
The population parameters of the NLME model \eqref{eq:NLME_model}-\eqref{eq:individual_parameter_matrix} are given by 
\begin{align*}
  \theta = (\zpop, \Omega, \sigma^2).
\end{align*}
The open parameter space of $\theta$ is denoted by $\Theta$,
\begin{align*}
  \Theta \;=\; \Reals^{n_z+n_\beta}\times\mathbb S^{n_z}_{++}\times\Reals_{>0},
\end{align*}
where $\mathbb S^{n_z}_{++}$ denotes the cone of symmetric positive definite $n_z\times n_z$ matrices.
The likelihood function of the observed data $x = (x_{ij}; \ 1\leq i \leq N,\, 1 \leq j \leq n_i)$ is given by  
\begin{align}
  \label{eq:marginalized_likelihood}
  p(x)=p(x; \theta) = \int p(x,z) \, dz = \int p(x|z) p(z) \, dz,
\end{align}
where
\begin{align*}
  p(x|z)= \prod_{i=1}^N \prod_{j=1}^{n_i} p(x_{ij}|z_i; \theta) \quad \text{and} \quad
  p(z) = \prod_{i=1}^N p(z_i; \theta).
\end{align*}
The corresponding log-likelihood is denoted by 
\begin{align*}
  \mathcal{LL}(\theta) = \mathcal{LL}(\theta; x) = \log p(x; \theta).
\end{align*}
The objective is to maximize the log-likelihood with respect to the population parameters
$\theta$, i.e.,
\begin{align}
  \label{eq:ML_problem}
  \max_{\theta} \mathcal{LL}(\theta),
\end{align}
referred to as marginalized likelihood problem.

\subsection{VAEs for NLME Models}

The integral \eqref{eq:marginalized_likelihood} also normalizes the posterior distribution of the
individual parameters \cite{Bishop2006}. VAEs replace this intractable posterior distribution by a
parametrized approximation $q(z|x)$
\cite{KingmaAutoEncodingVariationalBayes14,RezendeStochasticBackpropagationApproximate14},
\begin{align*}
    q(z|x)\;\approx\;p(z|x)\;=\;\frac{p(x|z)p(z)}{p(x)} .
\end{align*}
The encoder provides the variational approximation of the posterior, and the decoder is the NLME
model \eqref{eq:NLME_model}.

\subsubsection{Encoder}
\label{sec:encoder}

The encoder approximates the posterior $p(z_i|x_i)$ by a Gaussian distribution. It is a neural
network with parameters $\psi\in\Psi$, for NLME models typically a recurrent network
\cite{RohleffRedefiningParameter25}. It reads
\begin{align}
  \label{eq:encoder}
  \begin{aligned}
    (\mu_i, L_i) &= \mathrm{Encoder}_{\psi}(\mathcal{D}_i), \quad i = 1,\dots,N, \\
    q_\psi(z_i|x_i) &= \mathcal{N}(z_i;\mu_i,\Sigma_i), \quad \Sigma_i = L_i L_i^\top,
      \end{aligned}
\end{align}
where $\mathcal D_i=\big(x_i,(t_{ij})_{j=1}^{n_i},c_i\big)$ is the data of individual $i$ with
$x_i=(x_{ij})_{j=1}^{n_i}$, and $L_i$ is a lower triangular matrix with positive diagonal entries, so
that $\Sigma_i$ is positive definite. Both $\mu_i$ and $L_i$ depend on $\psi$.  
The individual parameters are sampled with the reparameterization trick
\cite{KingmaAutoEncodingVariationalBayes14},
\begin{align}
  \label{eq:reparameterization_trick}
    z_i = \mu_i + L_i \varepsilon_i, \quad \varepsilon_i \sim \mathcal{N}(0,I).
\end{align}
The reparameterization allows backpropagation through the sampling step, i.e., the encoder can be
trained by stochastic gradient-based optimization.

\subsubsection{Decoder}

The decoder is the structural model \eqref{eq:NLME_model}. For the sampled individual parameters
$z_i$ from \eqref{eq:reparameterization_trick}, the model prediction is
\begin{align*}
    \hat x_{ij}(z_i) = f(t_{ij}, h^{-1}(z_i)) \quad \text{for} \ i=1,\dots,N \text{ and } j=1,\dots,n_i.
\end{align*}

\subsubsection{The Evidence Lower Bound (ELBO)}

Since the integral in the marginalized likelihood problem \eqref{eq:ML_problem} is intractable, VAEs
maximize the ELBO
\begin{align*}
  \mathcal{L}_\psi(x;\theta)
  \;=\;\Exp_{z \sim q_\psi(z|x)} \!\left[\log \frac{p(x|z)p(z)}{q_\psi(z|x)}\right]
\end{align*}
instead. For every $\psi$, it holds
\cite{KingmaAutoEncodingVariationalBayes14,BleiVariationalInferenceReview17}
\begin{align}
  \label{eq:elbo_kl_decomposition}
\mathcal{LL}(x;\theta)
=
\mathcal{L}_\psi(x;\theta)
+
D_{KL}\!\big(q_\psi(z|x)\,\|\,p(z|x)\big)
\;\ge\;\mathcal{L}_\psi(x;\theta) ,
\end{align}
where $D_{KL}(q\|p)=\Exp_{z\sim q}\big[\log\big(q(z)/p(z)\big)\big]\ge0$ denotes the
Kullback-Leibler (KL) divergence. Hence, the ELBO is a lower bound of the log-likelihood, whose gap is the KL divergence between the
variational approximation and the posterior. As for the log-likelihood, the data are suppressed in the notation, i.e.,
$\mathcal{L}_\psi(\theta)=\mathcal{L}_\psi(x;\theta)$. The marginalized likelihood problem
\eqref{eq:ML_problem} is replaced by
\begin{align}
\label{eq:ELBO_optimization}
\max_{\theta,\psi} \mathcal{L}_\psi(x;\theta).
\end{align}
Separating the likelihood and the prior terms gives
\begin{align*}
\mathcal{L}_\psi(x;\theta)=\mathcal L_\psi(\theta)
=
\sum_{i=1}^N
\Big(
\Exp_{z_i \sim q_\psi(z_i|x_i)}[\log p(x_i|z_i)]
-
D_{KL}(q_\psi(z_i|x_i)\|p(z_i))
\Big).
\end{align*}
The likelihood term of the NLME model is
\begin{align*}
\log p(x_i|z_i)
=
-\frac{n_i}{2}\log(2\pi)
-\sum_{j=1}^{n_i}\log\!\big(g(t_{ij},h^{-1}(z_i))\sigma\big)
-\frac{1}{2\sigma^{2}}\sum_{j=1}^{n_i}
r_{ij}(z_i)^2,
\end{align*}
where
\begin{align*}
  r_{ij}(z_i) \;=\; \frac{x_{ij}-\hat x_{ij}(z_i)}{g(t_{ij},h^{-1}(z_i))}
\end{align*}
is the standardized residual. For Gaussian variational and prior distributions, the KL divergence
has the closed form
\begin{align}
  \label{eq:kl_closed_form}
D_{KL}(q_\psi(z_i|x_i)\|p(z_i))
&=
D_{KL}\big(\mathcal{N}(\mu_i,\Sigma_i)\,\|\,\mathcal{N}(C_i\zpop,\Omega)\big) \notag\\
&=
\frac{1}{2}\Big[
\tr(\Omega^{-1}\Sigma_i)
+(\mu_i-C_i\zpop)^\top \Omega^{-1}(\mu_i-C_i\zpop)
-n_z
+\log\frac{|\Omega|}{|\Sigma_i|}
\Big].
\end{align}

\subsubsection{Gradient and Monte Carlo Estimation of the ELBO}
Since the KL divergence has a closed form, the ELBO is estimated with the reparameterization
\eqref{eq:reparameterization_trick} and $M\in\mathbb{N}$ Monte Carlo samples by
\begin{align*}
\widehat{\mathcal L}_\psi(\theta; \varepsilon)
&= \frac{1}{M} \sum_{m=1}^M \sum_{i=1}^N 
\log p\big(x_i | \mu_i + L_i\varepsilon_i^{(m)}\big)
- \sum_{i=1}^N D_{KL}\big(q_\psi(z_i|x_i)\,\|\, p(z_i)\big),
\end{align*}
where $\varepsilon_i^{(m)}\sim\mathcal{N}(0,I)$ for $m=1,\dots,M$ are i.i.d. samples, collected in the
noise variable $\varepsilon=(\varepsilon_i^{(m)})_{i=1,\dots,N,\,m=1,\dots,M}$. Since the noise does
not depend on $\psi$, $\nabla_\psi\widehat{\mathcal L}_\psi(\theta;\varepsilon)$ is an unbiased estimator
of $\nabla_\psi\mathcal L_\psi(\theta)$ for a smooth model with bounded derivatives.

\subsubsection{Exponential Family Structure and Sufficient Statistics of the ELBO}

The distributions $p(x|z)$, $p(z)$ and $q_\psi(z|x)$ are Gaussian and belong to the exponential
family. Hence, the ELBO estimator $\widehat{\mathcal L}_\psi(\theta;\varepsilon)$ decomposes into
sufficient statistics and natural parameters,
\begin{align}
	\label{eq:exponential_family}
\widehat{\mathcal L}_\psi(\theta; \varepsilon)
= \big\langle S(x,\psi; \varepsilon), \zeta(\theta) \big\rangle -\kappa(\theta)
+ \mathcal H(\psi; \varepsilon) + \mathcal C .
\end{align}
The first two terms collect everything that depends on $\theta$. The sufficient statistics
\begin{align}
  \label{eq:sufficient_statistics}
  S(x,\psi;\varepsilon)=\big(S_1,\,S_2,\,S_3\big)
\end{align}
have the three blocks
\begin{align*}
  S_1 = \big(\mu_i\big)_{i=1}^N, \qquad
  S_2 = \sum_{i=1}^N \big(\mu_i\mu_i^\top + L_iL_i^\top\big), \qquad
  S_3 = \frac{1}{M}\sum_{m=1}^M\sum_{i=1}^N\sum_{j=1}^{n_i}
         r_{ij}\big(z_i(\varepsilon_i^{(m)})\big)^2 ,
\end{align*}
with $z_i(\varepsilon_i^{(m)}) = \mu_i + L_i\varepsilon_i^{(m)}$, the natural parameters are
\begin{align}
  \label{eq:natural_parameters}
\zeta(\theta) = \Big( (\Omega^{-1} C_i \zpop)_{i=1}^N,\ -\tfrac{1}{2} \Omega^{-1},\ -\tfrac{1}{2\sigma^{2}}  \Big),
\end{align}
and
\begin{align}
  \label{eq:kappa}
\kappa(\theta) =  \frac{N}{2} \log |\Omega|  + \frac{1}{2} \sum_{i=1}^N (C_i \zpop)^\top \Omega^{-1} (C_i \zpop) + \frac{\Ntot}{2} \log( \sigma^2),
\end{align}
where $\Ntot = \sum_{i=1}^N n_i$ denotes the total number of observations. The inner product is
understood componentwise,
$\langle S,\zeta\rangle=\sum_{i=1}^N\langle S_{1,i},\zeta_{1,i}\rangle+\tr(S_2^\top\zeta_2)+S_3\zeta_3$.
The entropy term collects all terms that depend on $\psi$ and $\varepsilon$ only, and $\mathcal C$
the remaining constants,
\begin{align*}
\mathcal H(\psi; \varepsilon) &= \sum_{i=1}^N \log \big|\det L_i\big|- \frac{1}{M} \sum_{m=1}^M\sum_{i=1}^{N} \sum_{j=1}^{n_i} \log\!\big(g(t_{ij}, h^{-1}(z_i(\varepsilon_i^{(m)})))\big),\\
\mathcal C &= -\frac{\Ntot}{2} \log(2\pi) + \frac{Nn_z}{2}.
\end{align*}
The sufficient statistics depend on $\varepsilon$ through the reparameterization
\eqref{eq:reparameterization_trick}. Averaging over the noise removes this dependence. Writing
\begin{align}
  \label{eq:averaged_quantities}
S(\psi) := \Exp_{\varepsilon \sim \mathcal N(0,I)}\big[S(x, \psi; \varepsilon)\big],
\qquad
\mathcal H(\psi) := \Exp_{\varepsilon \sim \mathcal N(0,I)}\big[\mathcal H(\psi; \varepsilon)\big],
\end{align}
the exact ELBO reads
\begin{align}
  \label{eq:elbo_expfam}
  \mathcal L_\psi(\theta) \;=\; J\big(\theta, S(\psi)\big) + \mathcal H(\psi) + \mathcal C,
  \qquad
  J(\theta,s) \;=\; \langle s,\zeta(\theta)\rangle-\kappa(\theta).
\end{align}

\section{The VAE-NLME Algorithm}
\label{sec:algorithm}
This section derives the VAE-NLME algorithm, starting from the ELBO problem
\eqref{eq:ELBO_optimization}. The algorithm alternates between an inner loop, which updates the
encoder, and an outer loop, which updates the population parameters. This structure follows the EM
algorithm \cite{DempsterMaximumLikelihoodIncomplete77,Bishop2006} and its stochastic approximation
version SAEM \cite{DelyonConvergenceStochasticApproximation99}. Importance-weighted statistics then target the
marginal log-likelihood instead of the ELBO and give the final algorithm.

\subsection{General Algorithmic Framework}
\label{sec:framework}
Let the initial parameters $\psi^{(0)}$ and $\theta^{(0)}$ be given. Iteration $k$ of the ELBO VAE-NLME
algorithm consists of three steps.
\begin{enumerate}
    \item \textbf{Simulation step:}
The ELBO is approximately maximized with respect to the encoder for fixed population parameters
$\theta^{(k-1)}$. The exact solution would be
\begin{align*}
\arg\max_{\psi\in\Psi} \mathcal L_\psi(\theta^{(k-1)}).
\end{align*}
It is approximated by $\nin$ stochastic gradient steps on the encoder parameters $\psi$,
\begin{align}
\label{eq:gradient_updates}
\psi_\ell
=
\psi_{\ell-1}
+
\alpha_k \,
\nabla_\psi \widehat{\mathcal L}_{\psi_{\ell-1}}\big(\theta^{(k-1)}; \varepsilon_\ell\big),
\quad \ell=1,\dots,\nin,
\end{align}
where $\varepsilon_\ell\sim\mathcal N(0,I)$ are i.i.d. samples, and $\psi^{(k)}=\psi_{\nin}$.
   \item \textbf{Stochastic approximation:} A noise sample $\varepsilon^{(k)}\sim\mathcal N(0,I)$ is drawn and $Q^{(k)}(\theta)$ is updated as
   \begin{align}
	\label{eq:stochastic_approximation}
	Q^{(k)}(\theta) =Q^{(k-1)}(\theta) + \gamma_k \big( \widehat{\mathcal L}_{\psi^{(k)}}(\theta; \varepsilon^{(k)}) - Q^{(k-1)}(\theta) \big).
   \end{align}

   \item \textbf{Maximization step (M-step):} The population parameters are updated as
   \begin{align}
	\label{eq:maximization_step}
	\theta^{(k)} = \arg\max_\theta Q^{(k)}(\theta).
   \end{align}
\end{enumerate}

\subsection{Exploiting the Exponential Family Structure}
\label{sec:mstep}

The recursion \eqref{eq:stochastic_approximation}-\eqref{eq:maximization_step} updates a function
of $\theta$. The exponential family structure reduces it to a recursion in the sufficient statistics,
whose dimension is denoted by $\dS=Nn_z+n_z(n_z+1)/2+1$. If the auxiliary function $Q^{(0)}$ is initialized in the form
$\langle s^{(0)},\zeta(\theta)\rangle-\kappa(\theta)+b^{(0)}$ with $s^{(0)}\in\Reals^{\dS}$ and a
scalar $b^{(0)}$, this form is preserved along the recursion \eqref{eq:stochastic_approximation},
\begin{align}
  \label{eq:Q_form}
  Q^{(k)}(\theta)\;=\;\big\langle s^{(k)},\zeta(\theta)\big\rangle-\kappa(\theta)+b^{(k)},
\end{align}
where the mean sufficient statistics and the scalar are updated by
\begin{align}
  \label{eq:sa_statistics}
  s^{(k)} &= s^{(k-1)} + \gamma_k \Big( S\big(x, \psi^{(k)}; \varepsilon^{(k)}\big) - s^{(k-1)} \Big), \\
  b^{(k)} &= b^{(k-1)} + \gamma_k \Big( \mathcal H\big(\psi^{(k)};\varepsilon^{(k)}\big)+\mathcal C - b^{(k-1)} \Big). \notag
\end{align}
The derivation is given in Appendix~\ref{app:reduction}.
Since $b^{(k)}$ does not depend on $\theta$, the M-step only requires $s^{(k)}$, and with $J$ from
\eqref{eq:elbo_expfam}, the M-step operator is defined by
\begin{align}
    \label{eq:m_operator}
    \mathcal{M}:\mathbb{R}^{\dS}\to\Theta,
    \quad
    \mathcal{M}(s)
    :=\arg\max_{\theta\in\Theta}J(\theta,s),
\end{align}
so that $\theta^{(k)}=\mathcal{M}(s^{(k)})$. As in the EM algorithm
\cite[Section~2]{DempsterMaximumLikelihoodIncomplete77}, $J(\cdot,s)$ is the log-likelihood of an
exponential family with sufficient statistics $s$. If this family is regular, e.g., for diagonal
$\Omega$, the maximizer is unique whenever $s$ lies in the interior of the convex support of the
sufficient statistics \cite[Theorem~5.5]{BrownFundamentalsStatisticalExponential86}.
\\
Furthermore, the statistics space is defined by
\begin{align}
    \label{eq:stat_space}
    \mathcal{S}
    &:=\overline{\operatorname{conv}}
    \left\{
        S(x,\psi;\varepsilon):
        \psi\in\Psi,\;
        \varepsilon\in\mathbb{R}^{NMn_z}
    \right\},
\end{align}
and the corresponding set of parameters attainable by the M-step,
\begin{align*}
    \Theta_{\mathcal{S}}
    &:=\mathcal{M}(\mathcal{S}).
\end{align*}

\subsection{The ELBO VAE-NLME Algorithm}
\label{sec:elbo_algorithm}

Instead of the function $Q^{(k)}(\theta)$, the algorithm updates the mean sufficient statistics.
Steps 2 and 3 of Section~\ref{sec:framework} read as follows.

\begin{itemize}
    \item[2.] \textbf{Stochastic approximation:} A noise sample $\varepsilon^{(k)}\sim\mathcal N(0,I)$
    is drawn, the sufficient statistics $S(x,\psi^{(k)};\varepsilon^{(k)})$ are computed, and the mean
    sufficient statistics are updated by
    \begin{align*}
        s^{(k)} = s^{(k-1)} + \gamma_k \Big( S\big(x, \psi^{(k)}; \varepsilon^{(k)}\big) - s^{(k-1)} \Big).
    \end{align*}

    \item[3.] \textbf{Maximization step:} The population parameters are updated with the M-step
    operator \eqref{eq:m_operator},
    \begin{align*}
        \theta^{(k)} = \mathcal M(s^{(k)}).
    \end{align*}
\end{itemize}
This algorithm is called the ELBO VAE-NLME algorithm, since it maximizes the ELBO \eqref{eq:ELBO_optimization}.

\subsection{Importance-Weighted Statistics}
\label{sec:is_statistics}

By \eqref{eq:elbo_kl_decomposition}, the ELBO maximized by the ELBO VAE-NLME algorithm differs from
the marginal log-likelihood by a KL divergence that depends on $\theta$. Hence, its maximum is in
general not the maximum likelihood estimator. The ELBO enters the M-step only through the
statistics. For $z=(z_i)_{i=1}^N$, the complete-data statistics are
\begin{align}
  \label{eq:complete_statistics}
  S(x,z) \;=\; \Big(\,(z_i)_{i=1}^N,\ \sum_{i=1}^N z_iz_i^\top,\
  \sum_{i=1}^N\sum_{j=1}^{n_i} r_{ij}(z_i)^2\Big),
\end{align}
i.e., $\log p(x,z;\theta)=\langle S(x,z),\zeta(\theta)\rangle-\kappa(\theta)+c(x,z)$ with $\zeta$,
$\kappa$ from \eqref{eq:natural_parameters}-\eqref{eq:kappa} and $c$ independent of $\theta$. The
averaged variational statistics $S(\psi)$ from \eqref{eq:averaged_quantities} are their expectation
under $q_\psi$. As in the EM algorithm and SAEM
\cite{DempsterMaximumLikelihoodIncomplete77,DelyonConvergenceStochasticApproximation99}, this
expectation is replaced by the expectation under the exact posterior distribution,
\begin{align*}
  \bar S_p(\theta)\;=\;\Exp_{z\sim p(\cdot|x;\theta)}\big[S(x,z)\big].
\end{align*}
The posterior statistics $\bar S_p(\theta)$ are estimated by importance sampling with the encoder as
proposal. In iteration $k$, $K$ samples $z_{i1},\dots,z_{iK}$ are drawn for every individual from the
multivariate $t$ distribution $r^{(k)}_i=t_\nu(\mu_i,L_iL_i^\top)$ with $\nu$ degrees of freedom and
the encoder output $(\mu_i,L_i)$ of $\psi^{(k)}$. The self-normalized weights and the weighted
statistics are
\begin{align}
  \label{eq:is_statistics}
  \begin{aligned}
  w_{ij}&=\frac{p(x_i,z_{ij};\theta^{(k-1)})}{r^{(k)}_i(z_{ij})},\qquad
  \tilde w_{ij}=\frac{w_{ij}}{\sum_{j'=1}^K w_{ij'}},\\[4pt]
  \widehat S_K&=\Big(\Big(\sum_{j=1}^K\tilde w_{ij}z_{ij}\Big)_{i=1}^N,\
  \sum_{i,j}\tilde w_{ij}\,z_{ij}z_{ij}^\top,\
  \sum_{i,j}\tilde w_{ij}\sum_{l=1}^{n_i} r_{il}(z_{ij})^2\Big).
  \end{aligned}
\end{align}
The $t$ proposal has heavier tails than the posterior distribution, which keeps the weights bounded
\cite{GewekeBayesianInferenceEconometric89,OwenMonteCarloTheory13}. In the stochastic approximation
of Section~\ref{sec:elbo_algorithm}, the weighted statistics $\widehat S_K$ replace the variational
statistics $S(x,\psi^{(k)};\varepsilon^{(k)})$. Simulation step and M-step are unchanged. The additional cost is
$K$ decoder evaluations per individual and iteration.

\subsection{The Final VAE-NLME Algorithm}
\label{sec:final_algorithm}

The final VAE-NLME algorithm combines the simulation step of Section~\ref{sec:framework} with the weighted
statistics of Section~\ref{sec:is_statistics}. Algorithm~\ref{alg:vae_nlme} summarizes it.

\begin{algorithm}[H]
\caption{VAE-NLME Algorithm}
\label{alg:vae_nlme}
\begin{algorithmic}[1]
\STATE \textbf{Input:} Initial parameters $\psi^{(0)}, \theta^{(0)}, s^{(0)}$, step sizes $\{\alpha_k\}, \{\gamma_k\}$, inner-loop length $\nin$, number $K$ of samples
\FOR{$k = 1, 2, \dots$}
    \STATE Set $\psi_0 = \psi^{(k-1)}$
    \STATE \textbf{Simulation step:}
    \FOR{$\ell = 1, \dots, \nin$}
        \STATE $\psi_\ell = \psi_{\ell-1} + \alpha_k \, \nabla_\psi \widehat{\mathcal L}_{\psi_{\ell-1}}(\theta^{(k-1)}; \varepsilon_\ell)$, where $\varepsilon_\ell \sim \mathcal N(0,I)$ i.i.d.
    \ENDFOR
    \STATE Set $\psi^{(k)} = \psi_{\nin}$
    \STATE \textbf{Stochastic Approximation:}
    \STATE Set $r^{(k)}_i=t_\nu(\mu_i,L_iL_i^\top)$ with $(\mu_i,L_i)=\mathrm{Encoder}_{\psi^{(k)}}(\mathcal D_i)$ for $i=1,\dots,N$
    \STATE Draw $z_{i1},\dots,z_{iK}\sim r^{(k)}_i$ for $i=1,\dots,N$ and compute $\widehat S_K$ by \eqref{eq:is_statistics}
	\STATE Update
    \[
    s^{(k)} = s^{(k-1)} + \gamma_k \big( \widehat S_K - s^{(k-1)} \big)
    \]
    \STATE \textbf{Maximization step:} $\theta^{(k)} = \mathcal M(s^{(k)})$
\ENDFOR
\end{algorithmic}
\end{algorithm}

\begin{remark}[Practical realization]
\label{rem:practice}
A run passes through the four phases (a)-(d), separated by two iteration counts
$K_{\mathrm{KL}}\le K_{\mathrm{sm}}$.
\begin{enumerate}
\item[(a)] \emph{Burn-in} ($k\le0$). The encoder is pre-trained with a small weight on the KL term,
      so that the first sufficient statistics already contain information about the data.
\item[(b)] \emph{KL annealing} ($1\le k<K_{\mathrm{KL}}$)
      \cite{BowmanGeneratingSentencesContinuous16,SonderbyLadderVariationalAutoencoders16}. The weight
      of the KL term \eqref{eq:kl_closed_form} is increased to one at $K_{\mathrm{KL}}$, so that the encoder
      first fits the data. Together with (a), this reduces the risk of poor local optima.
\item[(c)] \emph{Exploratory phase} ($K_{\mathrm{KL}}\le k\le K_{\mathrm{sm}}$). Here $\gamma_k=1$, i.e., $s^{(k)}$
      equals the statistics of the current iteration. This accelerates the convergence, while the
      population parameters fluctuate.
\item[(d)] \emph{Smoothing phase} ($k>K_{\mathrm{sm}}$). The step sizes $\gamma_k=1/(k-K_{\mathrm{sm}})$ satisfy
      the Robbins-Monro conditions \cite{RobbinsStochasticApproximationMethod51} and average out the
      fluctuations of (c), while $\alpha_k$ and $\nin$ stay fixed.
\item[(e)] \emph{Optimizer.} The gradient step \eqref{eq:gradient_updates} may be replaced by other
      optimizers, such as Adam.
\end{enumerate}
The counts $K_{\mathrm{KL}}$ and $K_{\mathrm{sm}}$ are finite and can be chosen adaptively.
\end{remark}

\section{Main Results}
\label{sec:convergence}

This section states the three main theorems.

\subsection{Variational Parameters and Assumptions}
\label{sec:assumptions}

\subsubsection{Variational parameters}

\begin{definition}[Variational parameters]
\label{def:vq}
The \emph{variational parameters} are
\begin{align*}
  \vq \;=\; (\vq_1,\dots,\vq_N),\qquad \vq_i=(\mu_i,L_i)\in\Reals^{n_z}\times\mathcal T_+,
\end{align*}
where $\mathcal T_+$ denotes the set of lower triangular $n_z\times n_z$ matrices with positive
diagonal, and the \emph{amortization map} realized by the encoder \eqref{eq:encoder} is
\begin{align*}
  \vq(\cdot):\ \Psi\to\Reals^{Nn_z}\times\mathcal T_+^N,
  \qquad
  \psi\mapsto \vq(\psi)=\big(\mathrm{Encoder}_\psi(\mathcal D_i)\big)_{i=1}^N .
\end{align*}
\end{definition}

\noindent
All quantities of the algorithm depend on the encoder only through its outputs $\vq(\psi)$,
\begin{align*}
  \mathcal L_\psi(\theta)=\mathcal L\big(\theta,\vq(\psi)\big),
  \qquad
  S(x,\psi;\varepsilon)=S\big(x,\vq(\psi);\varepsilon\big),
  \qquad
  \mathcal H(\psi;\varepsilon)=\mathcal H\big(\vq(\psi);\varepsilon\big).
\end{align*}
Hence, the assumptions below constrain the encoder outputs and not the network parameters $\psi$.
Since $\mathcal H$ and $\mathcal C$ in \eqref{eq:elbo_expfam} are independent of $\theta$, the
population parameters enter the ELBO only through the surrogate $J$,
\begin{align}
  \label{eq:grad_theta_elbo}
  \nabla_\theta \mathcal L(\theta,\vq) \;=\; \nabla_\theta J\big(\theta,S(\vq)\big)
  \qquad\text{for every }\vq .
\end{align}

\noindent
Throughout, $\|\cdot\|$ denotes the Euclidean norm. Matrices that are components of $\theta$, $s$ or
$\vq$ enter with their entries, i.e., $\|\theta\|^2=\|\zpop\|^2+\|\Omega\|_F^2+(\sigma^2)^2$ for
$\theta=(\zpop,\Omega,\sigma^2)$ with the Frobenius norm $\|\cdot\|_F$, and for linear maps,
$\|\cdot\|$ is the induced operator norm.

\subsubsection{Assumptions}

\begin{assumption}[Regularity]
\label{ass:regularity}
\begin{enumerate}
\item[(i)] \emph{Bounded model.} For all $i$ and $j$, $f$ is bounded and $g\ge g_{\min}>0$.
\item[(ii)] \emph{Smooth model.} For all $i$ and $j$, the two maps
      \begin{align*}
        z\mapsto f\big(t_{ij},h^{-1}(z)\big)
        \qquad\text{and}\qquad
        z\mapsto g\big(t_{ij},h^{-1}(z)\big)
      \end{align*}
      are three times continuously differentiable with bounded derivatives up to order three.
\item[(iii)] \emph{Variational parameters.} The encoder outputs take values in the compact convex set
      \begin{align*}
        \mathcal Q=\Big\{\vq=(\mu_i,L_i)_{i=1}^N:\ \|\mu_i\|\le\mu_{\max},\
        \ell_{\min}\le (L_i)_{aa}\le\ell_{\max},\ |(L_i)_{ab}|\le\ell_{\max}\ \Big\}
      \end{align*}
      for constants $0<\ell_{\min}\le\ell_{\max}<\infty$ and $\mu_{\max}<\infty$. In particular there
      are $0<\tau_{\min}\le\tau_{\max}<\infty$ with $\tau_{\min} I\preceq\Sigma_i\preceq\tau_{\max} I$
      for all $\vq\in\mathcal Q$ and all $i$. Since $\vq(\Psi)\subseteq\mathcal Q$, the
      statistics space \eqref{eq:stat_space} is from here on given by
      $\mathcal S=\overline{\mathrm{conv}}\{S(x,\vq;\varepsilon):\vq\in\mathcal Q,\ \varepsilon\in\Reals^{NMn_z}\}$.
\item[(iv)] \emph{Non-vanishing residuals.} At least one individual cannot be fitted exactly, i.e.
      \begin{align*}
        s_{\min}\;:=\;\sum_{i=1}^N\ \inf_{z\in\Reals^{n_z}}\ \sum_{j=1}^{n_i} r_{ij}(z)^2\;>\;0 .
      \end{align*}
\end{enumerate}
\end{assumption}

 \begin{assumption}[M-step local maximizer]
\label{ass:mstep}
There exists a continuously differentiable map $\mathcal M:\mathcal S\to\Theta$ such that, for every $s\in\mathcal S$,
\begin{align}
  \label{eq:m5}
  \nabla_\theta J\big(\mathcal M(s),s\big)&\;=\;0 , \\
  \label{eq:m5_hessian}
  \nabla^2_\theta J\big(\mathcal M(s),s\big)&\;\prec\;0.
\end{align}
\end{assumption}

\begin{assumption}[Variational local maximizer]
\label{ass:branch}
There are constants $\Kmin>0$, $r\in(0,\ell_{\min})$, an open set
$\mathcal O\supseteq\Theta_{\mathcal S}$ and a continuously differentiable map
\begin{align*}
  \vq^\star:\ \mathcal O\ \longrightarrow\ \mathcal Q,
  \qquad
  \theta\ \longmapsto\ \vq^\star(\theta)=\big(\mu^\star_i(\theta),L^\star_i(\theta)\big)_{i=1}^N ,
\end{align*}
such that for every $\theta\in\Theta_{\mathcal S}$
\begin{align}
  \label{eq:branch_foc}
  \nabla_\vq \mathcal L\big(\theta,\vq^\star(\theta)\big)&\;=\;0 , \\
  \label{eq:branch_hessian}
  \nabla^2_\vq \mathcal L\big(\theta,\vq\big)&\;\preceq\;-\Kmin I
  \qquad\text{for all }\vq\in\mathcal U(\theta),
\end{align}
where $\mathcal U(\theta)\;:=\;\big\{\vq:\ \big\|\vq-\vq^\star(\theta)\big\|\le r\big\}.
$
Define
\begin{align*}
  \mathcal L^\star(\theta)\;:=\;\mathcal L\big(\theta,\vq^\star(\theta)\big),
  \qquad
  \bar S(\theta)\;:=\;S\big(\vq^\star(\theta)\big).
\end{align*}
\end{assumption}

\begin{assumption}[Step sizes]
\label{ass:stepsizes}
The outer step sizes satisfy $\gamma_k\in(0,1]$, $\gamma_k$ non-increasing and
$\gamma_{k-1}/\gamma_k\to1$, together with the Robbins-Monro conditions
\cite{RobbinsStochasticApproximationMethod51}
\begin{align*}
    \sum_{k} \gamma_k = \infty, \qquad \sum_{k} \gamma_k^2 < \infty .
\end{align*}
\end{assumption}

\begin{assumption}[Inner loop]
\label{ass:contraction}
\begin{enumerate}
\item[(i)] \emph{Contraction.} There are $\rho\in[0,1)$ and $c_0<\infty$ such that, almost surely
      for every $k\ge1$, the encoder output $\vq^{(k)}=\vq(\psi^{(k)})$ of the simulation step has
      the inner-loop error
      \begin{align}
        \label{eq:contraction}
        \delta_k\;:=\;\big\|\vq^{(k)}-\vq^\star(\theta^{(k-1)})\big\|
        \;\le\;
        \rho\,\big\|\vq^{(k-1)}-\vq^\star(\theta^{(k-1)})\big\| \;+\; c_0\,\alpha_k .
      \end{align}
\item[(ii)] \emph{Coupled step sizes.} The inner step sizes satisfy $\alpha_k\le c_\alpha\gamma_k$ for
      some $c_\alpha<\infty$.
\end{enumerate}
\end{assumption}

\begin{assumption}[Compact iterates for the weighted statistics]
\label{ass:is}
The iterates $s^{(k)}$ of VAE-NLME (Algorithm~\ref{alg:vae_nlme}), i.e., with the weighted statistics
\eqref{eq:is_statistics}, remain almost surely in a compact convex set $\mathcal S_p\subset\Reals^{\dS}$ on which $s_3>0$ and
\begin{align*}
  \Sigtot(s)\;:=\;s_2-\sum_{i=1}^N s_{1,i}s_{1,i}^\top\;\succ\;0
  \qquad\text{for } s=(s_1,s_2,s_3),\ s_1=(s_{1,i})_{i=1}^N .
\end{align*}
Let $\Theta_{\mathcal S_p}:=\mathcal M(\mathcal S_p)$.
\end{assumption}

\subsection{Convergence to Stationary Points of the ELBO}
\label{sec:main_elbo}

The first two theorems concern the ELBO VAE-NLME algorithm of Section~\ref{sec:elbo_algorithm}.

\begin{theorem}[Convergence of the ELBO VAE-NLME algorithm]
\label{theo:convergence}
Let Assumptions~\ref{ass:regularity}-\ref{ass:contraction} hold, and
let $\{(s^{(k)},\theta^{(k)})\}$ be
generated by the ELBO VAE-NLME algorithm. Then the set
\begin{align*}
  \Theta^\star\;:=\;\big\{\theta\in\Theta_{\mathcal S}:\ \nabla_\theta\mathcal L^\star(\theta)=0\big\}
\end{align*}
is non-empty and compact, and, almost surely,
\begin{align*}
  \lim_{k\to\infty}\big\|\nabla_\theta\mathcal L^\star\big(\theta^{(k)}\big)\big\|=0
  \qquad\text{and}\qquad
  \lim_{k\to\infty}\ \min_{\theta^\star\in\Theta^\star}\big\|\theta^{(k)}-\theta^\star\big\|=0 .
\end{align*}
For every limit point $\theta^\star$ of $\{\theta^{(k)}\}$, the pair
$\big(\theta^\star,\vq^\star(\theta^\star)\big)$ is a stationary point of the ELBO
\eqref{eq:ELBO_optimization}. If in addition the elements of $\Theta^\star$ are isolated, then
$\theta^{(k)}$ converges almost surely to a single point $\theta^\star\in\Theta^\star$ and
$s^{(k)}\to\bar S(\theta^\star)$.
\end{theorem}

\noindent
If the inner loop does not contract, the inner-loop error does not necessarily vanish. In this case,
one gets a quantitative statement.

\begin{theorem}[Persistent inner-loop inaccuracy]
\label{theo:inexact}
Consider the ELBO VAE-NLME algorithm, let Assumptions~\ref{ass:regularity}-\ref{ass:stepsizes}
hold, and suppose that, instead of Assumption~\ref{ass:contraction}, the inner-loop error
$\delta_k$ from \eqref{eq:contraction} satisfies
\begin{align*}
  \limsup_{k\to\infty}\ \delta_k\;\le\;\delta
\end{align*}
for some $\delta\ge0$. Then there is a constant $C<\infty$, independent of $\delta$, such that,
almost surely, the following holds.
\begin{enumerate}
\item[(i)] \emph{Floor.}
      \begin{align}
        \label{eq:inexact_floor}
        \liminf_{k\to\infty}\ \big\|\nabla_\theta\mathcal L^\star\big(\theta^{(k)}\big)\big\|
        \;\le\; C\,\delta .
      \end{align}
\item[(ii)] \emph{Local convergence.} If $\theta^\star\in\Theta^\star$ is a local maximizer of
      $\mathcal L^\star$ with
      $m=\lambda_{\min}\big(-\nabla^2_\theta\mathcal L^\star(\theta^\star)\big)>0$, then there is a
      neighborhood $\mathcal U$ of $\theta^\star$ such that, for sufficiently small $\delta$, on
      the event that $\theta^{(k)}\in\mathcal U$ for infinitely many $k$,
      \begin{align}
        \label{eq:inexact_ball}
        \limsup_{k\to\infty}\ \big\|\theta^{(k)}-\theta^\star\big\|\;\le\;C_\delta,
        \qquad
        C_\delta\;:=\;\frac{2C}{m}\,\delta .
      \end{align}
\end{enumerate}
\end{theorem}

\begin{remark}[Fixed inner loop]
\label{rem:fixed_inner_loop}
Theorem~\ref{theo:convergence} does not cover the usual implementation of the algorithm. For a stochastic
inner loop ($c_0>0$) of fixed length $\nin$, the two parts of Assumption~\ref{ass:contraction} exclude
each other. For $\alpha_k\to0$, the contraction factor $(1-\alpha_k\Kmin)^{\nin}$ tends to one, and
for fixed $\alpha_k=\alpha$, the inner-loop error stabilizes at $c_0\alpha/(1-\rho)>0$. Both parts
together require $\nin\propto1/\gamma_k$. Theorem~\ref{theo:inexact} covers the fixed inner loop used
in practice and requires neither part.
\end{remark}

\subsection{Convergence to Stationary Points of the Log-Likelihood}
\label{sec:is}

The third theorem concerns VAE-NLME (Algorithm~\ref{alg:vae_nlme}).
Theorems~\ref{theo:convergence} and~\ref{theo:inexact} concern the ELBO, whereas the objective of
interest is the marginal log-likelihood \eqref{eq:ML_problem}. Since
\eqref{eq:elbo_kl_decomposition} holds for every $\vq$, evaluating it at $\vq^\star(\theta)$ gives
\begin{align}
  \label{eq:ll_decomposition}
  \mathcal{LL}(\theta)\;=\;\mathcal L^\star(\theta)\;+\;\mathcal G_{\mathrm{var}}(\theta),
  \qquad
  \mathcal G_{\mathrm{var}}(\theta):=\sum_{i=1}^N D_{KL}\Big(q_{\vq^\star_i(\theta)}\big(\cdot|x_i\big)\ \Big\|\ p\big(\cdot|x_i;\theta\big)\Big)\;\ge\;0 ,
\end{align}
and hence, at any limit point $\theta^\star$ of Theorem~\ref{theo:convergence},
\begin{align}
  \label{eq:ll_stationarity}
  \nabla_\theta\mathcal{LL}(\theta^\star)\;=\;\nabla_\theta\mathcal G_{\mathrm{var}}(\theta^\star).
\end{align}
If the Gaussian family contained the true posteriors, $\mathcal G_{\mathrm{var}}\equiv0$. The
weighted statistics \eqref{eq:is_statistics} remove the right-hand side of \eqref{eq:ll_stationarity},
without changing the encoder or the M-step. Their accuracy depends on how well the encoder
approximates the posterior distribution
\cite{AgapiouImportanceSamplingIntrinsic17,ChatterjeeSampleSizeRequired18}. Let $\theta_{\mathrm{ML}}$ denote a maximum likelihood estimator, i.e., a solution of
\eqref{eq:ML_problem}.

\begin{theorem}[Convergence of the VAE-NLME algorithm]
\label{theo:is}
Let Assumption~\ref{ass:regularity}(i) and~(iii), Assumption~\ref{ass:mstep} with $\mathcal S_p$
in place of $\mathcal S$, and Assumptions~\ref{ass:stepsizes} and~\ref{ass:is} hold, and let
$\{(s^{(k)},\theta^{(k)})\}$ be generated by VAE-NLME (Algorithm~\ref{alg:vae_nlme}). Then there is a constant
$C<\infty$, independent of $K$, which depends on the encoder only through the moments of the
weights along the iterates, such that, almost surely, the following holds.
\begin{enumerate}
\item[(i)] \emph{Floor.}
      \begin{align}
        \label{eq:is_floor}
        \liminf_{k\to\infty}\ \big\|\nabla_\theta\mathcal{LL}\big(\theta^{(k)}\big)\big\|
        \;\le\;\frac{C}{K} .
      \end{align}
\item[(ii)] \emph{Growing sample size.} If $K=K_k\to\infty$ with $\sum_k\gamma_k/K_k<\infty$, then
      $\nabla_\theta\mathcal{LL}(\theta^{(k)})\to0$, and $\theta^{(k)}$ converges to the set of
      stationary points of $\mathcal{LL}$ in $\Theta_{\mathcal S_p}$. If $\theta_{\mathrm{ML}}$ is
      the only stationary point there, then $\theta^{(k)}\to\theta_{\mathrm{ML}}$.
\item[(iii)] \emph{Local convergence.} If $\theta_{\mathrm{ML}}\in\Theta_{\mathcal S_p}$ satisfies
      $m=\lambda_{\min}\big(-\nabla^2_\theta\mathcal{LL}(\theta_{\mathrm{ML}})\big)>0$, then there
      is a neighborhood $\mathcal U$ of $\theta_{\mathrm{ML}}$ such that, on the event that
      $\theta^{(k)}\in\mathcal U$ for infinitely many $k$,
      \begin{align}
        \label{eq:is_ball}
        \limsup_{k\to\infty}\ \big\|\theta^{(k)}-\theta_{\mathrm{ML}}\big\|\;\le\;C_K,
        \qquad
        C_K\;:=\;\frac{2C}{m\,K},
      \end{align}
      if $C_K$ is sufficiently small, and $\theta^{(k)}\to\theta_{\mathrm{ML}}$ if $K=K_k\to\infty$.
\end{enumerate}
\end{theorem}

\noindent
In contrast to Theorems~\ref{theo:convergence} and~\ref{theo:inexact}, Theorem~\ref{theo:is}
requires no assumption on the training of the encoder, neither Assumption~\ref{ass:branch} nor
Assumption~\ref{ass:contraction}, and no smoothness of the model, i.e., no
Assumption~\ref{ass:regularity}(ii). The encoder only serves as proposal and affects the constant
$C$.

\subsection{Discussion}
\label{sec:discussion}

\subsubsection{The assumptions}
\label{sec:assumptions_discussion}

The assumptions are the usual conditions of the convergence analysis of stochastic approximation and
EM-type algorithms \cite{StochasticApproximationRecursive03,DelyonConvergenceStochasticApproximation99},
extended by conditions on the encoder. Each of them holds for typical NLME models or is
ensured by the implementation.
\begin{itemize}
\item \emph{Model.} Assumptions~\ref{ass:regularity}(i)-(ii) are naturally
satisfied by structural models defined through ODEs with smooth right-hand
sides, provided the individual parameters are restricted to bounded,
physiologically plausible ranges, as is standard in practical NLME modeling.
Part (iv) is naturally satisfied for noisy data, since no individual is
fitted exactly.
\item \emph{M-step.} Assumption~\ref{ass:mstep} is naturally satisfied
because $J(\cdot,s)$ corresponds to the log-likelihood of an exponential
family (Section~\ref{sec:mstep}). For regular exponential families, as
obtained, for example, with diagonal $\Omega$, the maximizer exists, is
unique, and depends smoothly on $s$ in the interior of the convex support
\cite[Theorem~5.5]{BrownFundamentalsStatisticalExponential86}.
\item \emph{E-step.} Assumption~\ref{ass:branch} imposes the analogous
regularity condition on the E-step. It requires a local maximizer of the
ELBO with respect to $\vq$ with a negative definite Hessian. This condition is naturally linked
to the local identifiability of the individual parameters. In
PMX, structural models are typically developed with
the aim of ensuring that individual parameters are identifiable from
the available data. Parameters that cannot be reliably identified are
usually removed or the model is otherwise simplified during model
development. Thus, Assumption~\ref{ass:branch} reflects a natural
desirable property of well-specified and identifiable NLME models.
\item \emph{Step sizes.} Assumption~\ref{ass:stepsizes} comprises the classical Robbins-Monro
conditions for stochastic approximation
\cite{RobbinsStochasticApproximationMethod51}, which are satisfied by
the chosen step-size sequence.
\\
     Assumption~\ref{ass:contraction} is the only assumption specifically
concerned with encoder training. Part (i) is satisfied by gradient
updates of the variational parameters under
Assumption~\ref{ass:branch}.
Part (ii) ensures that the inner loop tracks the moving target
$\vq^\star(\theta^{(k-1)})$, as required in two-timescale stochastic
approximation \cite{BorkarStochasticApproximationDynamical23}.
Importantly, without Assumption~\ref{ass:contraction},
Theorem~\ref{theo:inexact} remains applicable.
\item \emph{Stability.} Assumption~\ref{ass:is} is only needed for the weighted statistics. The
      variational statistics are convex combinations of bounded statistics and stay in
      $\mathcal S$ by construction. In contrast, $\widehat S_K$ is unbounded, and
      $\Sigtot(\widehat S_K)$ becomes singular if one weight dominates, i.e., if the weighted statistics
      collapse onto a single sample. As for SAEM
      \cite[Theorem~5]{DelyonConvergenceStochasticApproximation99},
      \cite{KuhnCouplingStochasticApproximation04}, the stability of the iterates is therefore
      assumed, and it can be enforced by reinitialization \cite{KuhnCouplingStochasticApproximation04}.
\end{itemize}

\subsubsection{The role of the encoder}
\label{sec:amortization}

The analysis is formulated in terms of the variational parameters
$\vq$ rather than the encoder parameters $\psi$. Indeed, by the chain
rule,
\begin{align*}
  \nabla_\psi\mathcal L\big(\theta,\vq(\psi)\big)
  =
  D\vq(\psi)^\top
  \nabla_\vq\mathcal L\big(\theta,\vq(\psi)\big).
\end{align*}
For overparameterized neural networks, $D\vq(\psi)$ is typically
rank-deficient. Consequently, stationarity with respect to $\psi$
does not necessarily imply stationarity with respect to $\vq$.
The encoder parameters therefore enter the convergence analysis
only through the approximation error $\delta_k$.
\\
To characterize this error, let $\mathcal L^\star_\Psi(\theta)$
denote the best ELBO achievable by the encoder, subject to
$\vq(\psi)\in\mathcal U(\theta)$. The difference
\begin{align*}
  \mathcal G_{\mathrm{am}}
  =
  \mathcal L^\star-\mathcal L^\star_\Psi
  \geq 0
\end{align*}
is the amortization gap, which quantifies the loss in ELBO due
to restricting the variational parameters to those representable
by the encoder. At $\theta=\theta^{(k-1)}$, the inference gap
can be decomposed into three components
\cite{CremerInferenceSuboptimalityVariational18},
\begin{align*}
\underbrace{\mathcal{LL}(\theta)-\mathcal L\big(\theta,\vq^{(k)}\big)}_{\text{inference gap}}
\;&=\;
\underbrace{\mathcal{LL}(\theta)-\mathcal L^\star(\theta)}_{\text{(a) variational family gap}}
\;+\;
\underbrace{\mathcal G_{\mathrm{am}}(\theta)}_{\text{(b) amortization gap}}
\;+\;
\underbrace{\mathcal L^\star_\Psi(\theta)-\mathcal L\big(\theta,\vq^{(k)}\big)}_{\text{(c) optimization gap}} \\
&= \mathcal G_{\mathrm{var}}(\theta) + \mathcal G_{\mathrm{am}}(\theta) + \mathcal G_{\mathrm{opt}}(\theta) .
\end{align*}
Part (a) is the variational gap of \eqref{eq:ll_decomposition}, the approximation error of the
Gaussian family, and does not affect Theorem~\ref{theo:convergence}. Parts (b) and (c) are measured
by $\delta_k$. Once $\vq^{(k)}\in\mathcal U(\theta^{(k-1)})$, a Taylor expansion around
$\vq^\star(\theta^{(k-1)})$ gives, with $\Kmin$ from Assumption~\ref{ass:branch} and the bound
$\Kmax<\infty$ of $\|\nabla_\vq^2\mathcal L\|$ on $\mathcal Q$,
\begin{align}
\label{eq:gap_delta}
\frac{\Kmin}{2}\,\delta_k^2
\;\le\;
\mathcal L^\star\big(\theta^{(k-1)}\big)-\mathcal L\big(\theta^{(k-1)},\vq^{(k)}\big)
\;\le\;
\frac{\Kmax}{2}\,\delta_k^2 .
\end{align}
Hence, a persistent amortization gap contributes to the residual in
Theorem~\ref{theo:inexact} at the order of
$\sqrt{\mathcal G_{\mathrm{am}}+\mathcal G_{\mathrm{opt}}}$, where
$\mathcal G_{\mathrm{opt}}$ denotes the limit superior of the
optimization gap in (c). If the inner optimization problem is solved
exactly with respect to $\psi$, the residual is reduced to
$\sqrt{\mathcal G_{\mathrm{am}}}$.

\subsubsection{Relation to SAEM}
\label{sec:saem}

The VAE-NLME algorithm is an SAEM algorithm
\cite{DelyonConvergenceStochasticApproximation99} with the same statistics, M-step and Lyapunov
function $-\mathcal{LL}\circ\mathcal M$. Only the simulation step differs. SAEM samples the
individual parameters from the exact posterior distribution by MCMC
\cite{KuhnCouplingStochasticApproximation04,KuhnMaximumLikelihoodEstimation05}, whereas the VAE-NLME
algorithm draws them from the encoder and weights them. Hence, the simulation step is
differentiable, and neural network components of the model are trained in the same iteration.

\section{Proof Strategy}
\label{sec:proofs}

\begin{figure}[!htb]
    \centering
    \includegraphics[width=\textwidth]{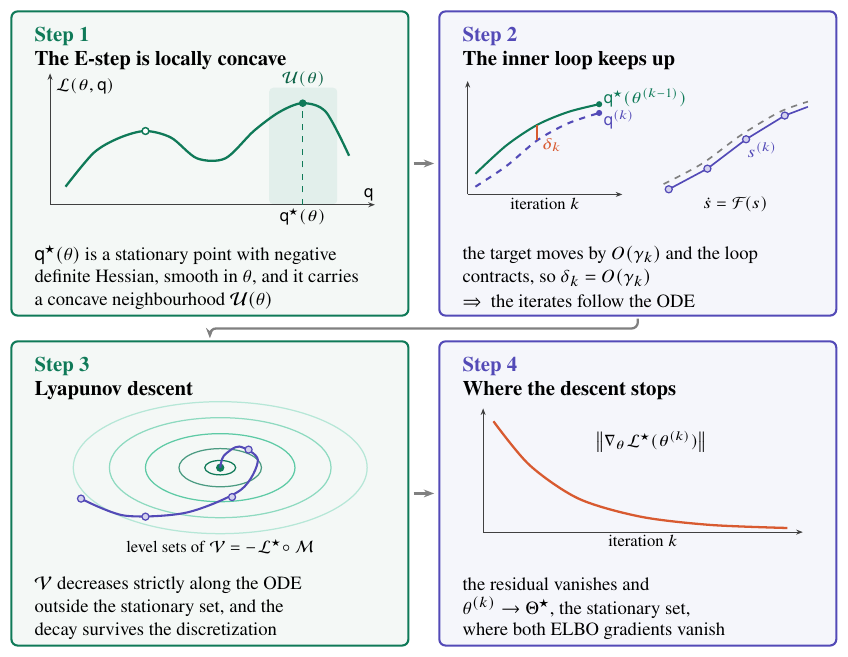}
    \caption{The four steps of the convergence proof.}
    \label{fig:proof_idea}
\end{figure}

The proof of Theorems~\ref{theo:convergence} and~\ref{theo:inexact} follows the two parts of the
algorithm, the inner loop and the outer update. Figure~\ref{fig:proof_idea} gives an overview of the
four steps.

\noindent
Theorem~\ref{theo:is} follows from the same steps, with the importance sampling bias in place of the
inner-loop error. All proofs are given in the Appendix, those of the three theorems in
Appendix~\ref{app:main}.

\subsection{Preliminaries: Statistics Space and M-Step Operator}
\label{sec:prelim}

\begin{stepbox}[stepgrey]{Preliminaries}
\textit{Idea.} The algorithm only stores the averaged sufficient statistics $s^{(k)}$, and the
population parameters follow from them by the M-step, $\theta^{(k)}=\mathcal M(s^{(k)})$. Hence, the
algorithm is studied in the space of the statistics. Two properties are needed.
\begin{enumerate}
  \item The statistics stay in a bounded set $\mathcal S$.
  \item The M-step reacts smoothly to a change of the statistics.
\end{enumerate}
\textit{Uses.} A regular model and an M-step with a local maximizer
(Assumptions~\ref{ass:regularity} and~\ref{ass:mstep}).
\\[2pt]
\textit{Provides.} Bounded iterates without further assumptions, since every update is a convex combination,
and a Lipschitz M-step with explicit derivative. All four steps use both.
\end{stepbox}

\begin{lemma}[The statistics space]
\label{lemma:statspace}
Let Assumption~\ref{ass:regularity} hold. Then $\mathcal S$ is compact and convex,
$s_3\ge s_{\min}>0$ on $\mathcal S$, and
\begin{align*}
  \Sigtot(s)\;\succeq\;N\tau_{\min} I\;\succ\;0 \qquad\text{for every } s\in\mathcal S .
\end{align*}
Moreover, if $\gamma_k\in[0,1]$ for all $k$ and $s^{(0)}\in\mathcal S$, then the iterates of the
ELBO VAE-NLME algorithm satisfy $s^{(k)}\in\mathcal S$ for all $k$ almost surely.
\end{lemma}

\noindent
The last statement is important, since keeping the iterates in a compact subset of the statistics space is
the hardest condition to verify in stochastic approximation. Here it holds by construction, since
$\gamma_k\in[0,1]$ makes \eqref{eq:sa_statistics} a convex combination.

\begin{lemma}[Properties of the M-step operator]
\label{lemma:mop}
Let Assumptions~\ref{ass:regularity} and~\ref{ass:mstep} hold, and write $\theta_s=\mathcal M(s)$. Then
\begin{enumerate}
\item[(i)] the matrix $H_s:=-\nabla_\theta^2 J(\theta_s,s)$ is positive definite for every
      $s\in\mathcal S$, and $\sup_{s\in\mathcal S}\|H_s\|<\infty$ as well as
      $\sup_{s\in\mathcal S}\|H_s^{-1}\|<\infty$,
\item[(ii)] $\Theta_{\mathcal S}=\mathcal M(\mathcal S)$ is a compact subset of $\Theta$, and
      $\mathcal M$ is Lipschitz continuous on $\mathcal S$,
\item[(iii)] the M-step reacts to a change of the statistics through
      \begin{align}
        \label{eq:mstep_sens}
        \nabla_s\mathcal M(s) \;=\; H_s^{-1}\,\Dzeta(\theta_s)^\top ,
      \end{align}
      where $\Dzeta$ is the Jacobian of the natural parameter map $\zeta$ from
      \eqref{eq:natural_parameters}.
\end{enumerate}
\end{lemma}

\subsection{Step 1: The Variational E-Step}
\label{sec:step1}

\begin{stepbox}[stepgreen]{Step 1}
\textit{Idea.} For fixed population parameters $\theta$, the encoder solves the E-step, i.e., it
maximizes the ELBO over its outputs $\vq$. This problem has a local solution
$\vq^\star(\theta)$, which changes smoothly with $\theta$. Hence, the ELBO with the best encoder
output, $\mathcal L^\star(\theta)=\mathcal L(\theta,\vq^\star(\theta))$, is a smooth function of
$\theta$ alone, and its gradient has the simple form of the M-step objective.
\\[2pt]
\textit{Uses.} A regular model and a locally concave E-step
(Assumptions~\ref{ass:regularity}-\ref{ass:branch}).
\\[2pt]
\textit{Provides.} The target $\vq^\star(\theta)$ of the inner loop (Step~2), the objective
$\mathcal L^\star$ that the algorithm increases (Steps~3 and~4), and the envelope identity.
\end{stepbox}

In the variational parameters, the E-step splits over the individuals,
$\mathcal L(\theta,\vq)=\sum_{i=1}^N\mathcal L_i(\theta,\vq_i)$ with
\begin{align}
  \label{eq:elbo_individual}
  \mathcal L_i(\theta,\vq_i)
  =\Exp_{\varepsilon\sim\mathcal N(0,I)}\big[\log p(x_i,\mu_i+L_i\varepsilon;\theta)\big]
  +\log\big|\det L_i\big|+\tfrac{n_z}{2}\log(2\pi e),
\end{align}
and the reparameterization \eqref{eq:reparameterization_trick} is affine in $\vq_i=(\mu_i,L_i)$.
Hence, Assumption~\ref{ass:branch} is a condition on the model.

\begin{lemma}[Smoothness and the maximizer on $\mathcal U(\theta)$]
\label{lemma:tube}
Let Assumptions~\ref{ass:regularity}-\ref{ass:branch} hold. Then $\vq\mapsto\mathcal L(\theta,\vq)$
is $\Kmax$-smooth on $\mathcal Q$, i.e.
\begin{align*}
  \big\|\nabla_\vq \mathcal L(\theta,\vq_1)-\nabla_\vq \mathcal L(\theta,\vq_2)\big\|
  &\;\le\;\Kmax\,\big\|\vq_1-\vq_2\big\|
  \qquad\text{for all }\vq_1,\vq_2\in\mathcal Q,
\end{align*}
with $\Kmax<\infty$ depending only on the bounds of
Assumption~\ref{ass:regularity} and on the compact set $\Theta_{\mathcal S}$, and for every
$\theta\in\Theta_{\mathcal S}$ the point
$\vq^\star(\theta)$ is the unique maximizer of $\mathcal L(\theta,\cdot)$ on $\mathcal U(\theta)$.
\end{lemma}

\begin{lemma}[The variational envelope identity]
\label{lemma:envelope}
Let Assumptions~\ref{ass:regularity}-\ref{ass:branch} hold. Then $\vq^\star$ and
$\bar S=S\circ\vq^\star$ are Lipschitz on $\Theta_{\mathcal S}$, $\bar S$ takes values in $\mathcal S$,
and $\mathcal L^\star$ is continuously differentiable with a Lipschitz gradient and
\begin{align}
  \label{eq:envelope}
  \nabla_\theta \mathcal L^\star(\theta)
  \;=\;\nabla_\theta\mathcal L\big(\theta,\vq^\star(\theta)\big)
  \;=\;\nabla_\theta J\big(\theta,\bar S(\theta)\big)
  \;=\;\Dzeta(\theta)^\top \bar S(\theta)-\nabla_\theta\kappa(\theta) .
\end{align}
\end{lemma}

\subsection{Step 2: Robbins-Monro Form, Inner-Loop Error and Limiting ODE}
\label{sec:step2}

\begin{stepbox}[steppurple]{Step 2}
\textit{Idea.} One iteration changes the sufficient statistics only by a small step of size $\gamma_k$. Hence,
the iterates behave like the solution of an ODE,
$\dot s=\Fmf(s)$, perturbed by Monte Carlo noise and by the error of the inner loop. Both
perturbations vanish on average, so that the iterates follow the ODE for large $k$.
\\[2pt]
\textit{Uses.} The target $\vq^\star(\theta)$ and the statistics $\bar S$ from Step~1, the Lipschitz
M-step from the preliminaries, the step sizes (Assumption~\ref{ass:stepsizes}) and the contracting
inner loop (Assumption~\ref{ass:contraction}).
\\[2pt]
\textit{Provides.} The mean field $\Fmf$ and the statement that the iterates follow its ODE. Hence, it suffices to study the ODE in Step~3.
\end{stepbox}

\subsubsection{The Robbins-Monro decomposition}

Let $\theta^{(k-1)}=\mathcal M(s^{(k-1)})$, and define the mean field
\begin{align*}
  \Fmf:\mathcal S\to\Reals^{\dS},\qquad \Fmf(s)\;=\;\bar S\big(\mathcal M(s)\big)-s .
\end{align*}
By Lemmas~\ref{lemma:mop} and~\ref{lemma:envelope} the mean field is Lipschitz and bounded on
$\mathcal S$. The update \eqref{eq:sa_statistics}
takes the exact Robbins-Monro form
\begin{align}
  \label{eq:rm_form}
  s^{(k)} \;=\; s^{(k-1)}+\gamma_k\Big(\Fmf\big(s^{(k-1)}\big)+\Delta^{\mathrm{mc}}_k+\Delta^{\mathrm{in}}_k\Big),
\end{align}
with
\begin{align*}
  \Delta^{\mathrm{mc}}_k := S\big(x,\vq^{(k)};\varepsilon^{(k)}\big)-S\big(\vq^{(k)}\big),
  \qquad
  \Delta^{\mathrm{in}}_k := S\big(\vq^{(k)}\big)-\bar S\big(\theta^{(k-1)}\big),
\end{align*}
where $\Delta^{\mathrm{mc}}_k$ is the Monte Carlo noise and $\Delta^{\mathrm{in}}_k$ is the inner-loop error. The next lemma collects the properties of these two terms that are needed for the convergence proof.

\begin{lemma}[Noise conditions]
\label{lemma:noise}
Let Assumptions~\ref{ass:regularity}-\ref{ass:stepsizes} hold and let $\mathcal G_k$ be the
$\sigma$-algebra generated by all random variables up to and including the $k$-th inner loop. Then
$\Exp[\Delta^{\mathrm{mc}}_k\mid\mathcal G_k]=0$, the Monte Carlo term is bounded by $2C_S$ with
$C_S=\sup_{\vq\in\mathcal Q,\varepsilon}\|S(x,\vq;\varepsilon)\|<\infty$, the series
$\sum_{k}\gamma_k\Delta^{\mathrm{mc}}_k$ converges almost surely, and
\begin{align}
  \label{eq:delta_in_bound}
  \big\|\Delta^{\mathrm{in}}_k\big\|\;\le\; \Lip{S}\,\delta_k
\end{align}
for the Lipschitz constant $\Lip{S}$ of $\vq\mapsto S(\vq)$ on $\mathcal Q$.
\end{lemma}

\subsubsection{The inner-loop error vanishes}
\begin{proposition}[Vanishing inner-loop error]
\label{prop:tracking}
Let Assumptions~\ref{ass:regularity}-\ref{ass:contraction} hold. Then, almost surely,
\begin{align*}
  \delta_k \;=\; O(\gamma_k)\;\xrightarrow[k\to\infty]{}\;0
  \qquad\text{and}\qquad
  \sum_{k\ge1}\gamma_k\,\delta_k \;<\;\infty .
\end{align*}
\end{proposition}

\subsubsection{The limiting ODE}

Define the interpolation times
\begin{align*}
t_0=0,
\quad
t_k=\sum_{j=1}^k\gamma_j,
\end{align*}
and the interpolated process
\begin{align*}
\bar s(t_k)=s^{(k)},
\quad
\bar s(t)
=
s^{(k)}
+
\frac{t-t_k}{t_{k+1}-t_k}
\big(s^{(k+1)}-s^{(k)}\big) \text{ for }
t\in[t_k,t_{k+1}].
\end{align*}
and let $\Flow$ denote the semiflow of the autonomous ODE
\begin{align}
    \label{eq:ode_final}
\dot{s}(t) \;=\; \Fmf(s(t)) \;=\; \bar{S}\big(\mathcal{M}(s(t))\big) - s(t).
\end{align}
Since $\Fmf$ is Lipschitz on $\mathcal S$, \eqref{eq:ode_final} has a unique solution for every initial
value \cite{NonlinearSystems3rd}. As $\mathcal S$ is convex and
$\bar S(\Theta_{\mathcal S})\subseteq\mathcal S$, the field points into $\mathcal S$ on
$\partial\mathcal S$, so $\mathcal S$ is forward invariant and $\Flow_v(s)$ is defined for all $v\ge0$
and all $s\in\mathcal S$.

\begin{proposition}[The interpolated process is an asymptotic pseudo-trajectory]
\label{prop:apt}
Let Assumptions~\ref{ass:regularity}-\ref{ass:contraction} hold. Then,
almost surely, for every $T>0$
\begin{align}
  \label{eq:apt}
  \lim_{t\to\infty}\ \sup_{0\le v\le T}\ \big\|\bar s(t+v)-\Flow_v\big(\bar s(t)\big)\big\| \;=\;0 ,
\end{align}
and the set of limit points of $\{s^{(k)}\}$ is a non-empty, compact, connected and invariant subset of
$\mathcal S$ for the flow of \eqref{eq:ode_final}.
\end{proposition}

\subsection{Step 3: Lyapunov Stability}
\label{sec:step3}

\begin{stepbox}[stepgreen]{Step 3}
\textit{Idea.} Along the ODE of Step~2, the ELBO $\mathcal L^\star$ increases, i.e.,
$\Lyap=-\mathcal L^\star\circ\mathcal M$ decreases. The
decrease is strict unless the gradient of $\mathcal L^\star$ vanishes. Since the iterates follow the
ODE, the algorithm shows the same decrease, up to the vanishing perturbations.
\\[2pt]
\textit{Uses.} The gradient of $\mathcal L^\star$ from Step~1, the derivative of the M-step from the
preliminaries, and the ODE with its perturbations from Step~2.
\\[2pt]
\textit{Provides.} The gradient of $\mathcal L^\star$ vanishes along the iterates.
\end{stepbox}
The Lyapunov function is defined by
\begin{align*}
\Lyap:\mathcal S\to\Reals,\qquad
\Lyap(s) \;=\; -\,\mathcal L^\star\big(\mathcal M(s)\big),
\qquad \theta_s=\mathcal M(s),
\end{align*}
which by Lemmas~\ref{lemma:mop} and~\ref{lemma:envelope} is bounded and continuously
differentiable on $\mathcal S$ with a Lipschitz gradient (Appendix~\ref{app:step3}).

\begin{lemma}[Lyapunov decay]
\label{lemma:lyapunov}
Let Assumptions~\ref{ass:regularity}-\ref{ass:branch} hold. Then, for every $s\in\mathcal S$,
\begin{align}
  \label{eq:decay}
  \big\langle \nabla_s\Lyap(s),\,\Fmf(s)\big\rangle
  \;=\;-\,\big\|\nabla_\theta\mathcal L^\star(\theta_s)\big\|^2_{H_s^{-1}}\;\le\;0 ,
\end{align}
and there are constants $\cdec,\cgrad\in(0,\infty)$ with
\begin{align}
  \label{eq:F_bounds}
  \big\langle \nabla_s\Lyap(s),\,\Fmf(s)\big\rangle\;\le\; -\cdec\,\big\|\nabla_\theta\mathcal L^\star(\theta_s)\big\|^2,
  \qquad
  \big\|\nabla_s\Lyap(s)\big\|\;\le\; \cgrad\,\big\|\nabla_\theta\mathcal L^\star(\theta_s)\big\| .
\end{align}
In particular $\Lyap$ decreases strictly along \eqref{eq:ode_final} outside the stationary set
\begin{align*}
  \mathcal S^\star\;:=\;\big\{s\in\mathcal S:\ \nabla_\theta\mathcal L^\star(\mathcal M(s))=0\big\} .
\end{align*}
\end{lemma}

\begin{proposition}[Discrete Lyapunov decay]
\label{prop:discrete_decay}
Let Assumptions~\ref{ass:regularity}-\ref{ass:contraction} hold. Then,
almost surely, $\Lyap(s^{(k)})$ converges to a finite limit and
\begin{align}
  \label{eq:V_converges}
  \sum_{k\ge1}\gamma_k\,\big\|\nabla_\theta\mathcal L^\star\big(\theta^{(k-1)}\big)\big\|^2 \;<\;\infty .
\end{align}
\end{proposition}

\begin{proposition}[Vanishing stationarity residual]
\label{prop:F_to_zero}
Under the assumptions of Proposition~\ref{prop:discrete_decay}, almost surely
\begin{align*}
  \lim_{k\to\infty}\big\|\nabla_\theta \mathcal L^\star\big(\theta^{(k)}\big)\big\|\;=\;0 .
\end{align*}
\end{proposition}

\subsection{Step 4: Stationary Optimality Conditions}
\label{sec:step4}

Step~3 is a statement about $\mathcal L^\star$, i.e., about $\theta$ alone. The last step translates
it back to the ELBO, which depends on $\theta$ and on the encoder outputs $\vq$.

\begin{lemma}[Characterization of the limit set]
\label{lemma:limitset}
Let Assumptions~\ref{ass:regularity}-\ref{ass:branch} hold and let
$\theta^\star\in\Theta_{\mathcal S}$ satisfy $\nabla_\theta\mathcal L^\star(\theta^\star)=0$. Put
$\vq^\star=\vq^\star(\theta^\star)$ and $s^\star=\bar S(\theta^\star)$. Then
\begin{align}
  \label{eq:limit_characterization}
  \nabla_\theta\mathcal L(\theta^\star,\vq^\star)=0,
  \qquad
  \nabla_\vq\mathcal L(\theta^\star,\vq^\star)=0,
  \qquad
  \nabla_\theta J\big(\theta^\star,s^\star\big)=0 .
\end{align}
The first two identities say that $(\theta^\star,\vq^\star)$ is a stationary point of the ELBO
\eqref{eq:ELBO_optimization}, the third that $\theta^\star$ solves the M-step at $s^\star$, so that the
algorithm does not move away from $(s^\star,\theta^\star)$. Conversely, every pair with these three
properties, with $\vq^\star=\vq^\star(\theta^\star)$ from Assumption~\ref{ass:branch}, satisfies
$\nabla_\theta\mathcal L^\star(\theta^\star)=0$.
\end{lemma}
\noindent
Proposition~\ref{prop:F_to_zero} and Lemma~\ref{lemma:limitset} give
Theorem~\ref{theo:convergence}. For Theorem~\ref{theo:inexact}, the inner-loop error does not vanish.
By \eqref{eq:delta_in_bound}, it enters the decay of Step~3 as a perturbation of size
$\Lip{S}\delta_k$, which leaves the floor \eqref{eq:inexact_floor} with $C=\cgrad\Lip{S}/\cdec$.

\subsection{Extension to Importance-Weighted Statistics}
\label{sec:is_proof}

The proof of Theorem~\ref{theo:is} follows the same four steps with the exact posterior distribution
instead of the variational distribution. This is the convergence
proof of SAEM \cite[Theorem~5]{DelyonConvergenceStochasticApproximation99}, only with biased and
unbounded noise.
\\
 The weighted statistics remove the right-hand side of \eqref{eq:ll_stationarity} because of the
Fisher identity \cite{DempsterMaximumLikelihoodIncomplete77,DelyonConvergenceStochasticApproximation99}
\begin{align}
  \label{eq:fisher}
  \nabla_\theta\mathcal{LL}(\theta)
  \;=\;\Dzeta(\theta)^\top\bar S_p(\theta)-\nabla_\theta\kappa(\theta)
  \;=\;\nabla_\theta J\big(\theta,\bar S_p(\theta)\big) .
\end{align}
This is the envelope identity \eqref{eq:envelope} with $\mathcal{LL}$ and $\bar S_p$ in place of
$\mathcal L^\star$ and $\bar S$. Hence, if the statistics are driven by $\bar S_p$, a fixed point
$s=\bar S_p(\mathcal M(s))$ satisfies $$\nabla_\theta\mathcal{LL}(\mathcal M(s))=\nabla_\theta
J(\mathcal M(s),s)=0$$ by \eqref{eq:m5}, i.e., the fixed points of the algorithm are stationary
points of the log-likelihood.
\\
With $\bar S_p$ in place of $\bar S$, the mean field and the Lyapunov function of Step~3 read
$\Fmf_p(s)=\bar S_p(\mathcal M(s))-s$ and $\Lyap_p(s)=-\mathcal{LL}(\mathcal M(s))$. The update is
again of the Robbins-Monro form
\eqref{eq:rm_form},
\begin{align*}
  s^{(k)} = s^{(k-1)}+\gamma_k\Big(\Fmf_p\big(s^{(k-1)}\big)+\Delta^{\mathrm{mc}}_k+\Delta^{\mathrm{is}}_k\Big),
  \qquad
  \Delta^{\mathrm{is}}_k:=\Exp\big[\widehat S_K\,\big|\,\mathcal G_k\big]-\bar S_p\big(\theta^{(k-1)}\big),
\end{align*}
with the Monte Carlo noise $\Delta^{\mathrm{mc}}_k=\widehat S_K-\Exp[\widehat S_K|\mathcal G_k]$.
The importance sampling bias $\Delta^{\mathrm{is}}_k$ replaces the inner-loop error. Its size depends
on the moments of the weights, in particular on the second moment
\cite{AgapiouImportanceSamplingIntrinsic17,ChatterjeeSampleSizeRequired18},
\begin{align*}
  \rho(\theta,\vq)\;=\;\max_{i\le N}\ \Exp_{z\sim r_i}\Big[\Big(\frac{p(z|x_i;\theta)}{r_i(z)}\Big)^{2}\Big]
  \;=\;1+\max_{i\le N}\chi^2\Big(p(\cdot|x_i;\theta)\,\Big\|\,r_i\Big),
\end{align*}
with the chi-square divergence $\chi^2(p\|r)=\int\big(p(z)-r(z)\big)^2/r(z)\,dz$. Hence, $\rho\ge1$
measures how well the encoder approximates the posterior distribution, with $\rho=1$ if both
coincide. The following lemma controls the bias.

\begin{lemma}[Consistency of the weighted statistics]
\label{lemma:is_consistent}
Let Assumption~\ref{ass:regularity}(i) hold, let $\theta\in\Theta$, $\vq\in\mathcal Q$, and let
$\widehat S_K$ be given by \eqref{eq:is_statistics} with $\nu>4$. Then $\widehat S_K\to\bar
S_p(\theta)$ almost surely as $K\to\infty$, and for every finite $K$,
\begin{align}
  \label{eq:is_finite_K}
  \big\|\Exp[\widehat S_K]-\bar S_p(\theta)\big\|\;\le\;\frac{c_{\mathrm{is}}(\theta,\vq)}{K},
  \qquad
  \Exp\Big[\big\|\widehat S_K-\bar S_p(\theta)\big\|^2\Big]\;\le\;\frac{c_{\mathrm{is}}(\theta,\vq)}{K},
\end{align}
where $c_{\mathrm{is}}(\theta,\vq)<\infty$ is independent of $K$, bounded on compact subsets of
$\Theta\times\mathcal Q$, and depends on the encoder only through the moments of the weights. For
the bounded third block of $\widehat S_K$, it is proportional to $\rho(\theta,\vq)$.
\end{lemma}
\noindent
Hence, 
\begin{align*}
  \|\Delta^{\mathrm{is}}_k\|\le \frac{\bar c_{\mathrm{is}}}{K} \quad
\text{with} \quad \bar c_{\mathrm{is}}=\sup\{c_{\mathrm{is}}(\theta,\vq):\theta\in\Theta_{\mathcal S_p},\
\vq\in\mathcal Q\}<\infty.
\end{align*}
Therefore, the proof of Theorem~\ref{theo:inexact} applies with
$\Lip{S}\delta$ replaced by $\bar c_{\mathrm{is}}/K$, i.e., with $C=\cgrad\bar c_{\mathrm{is}}/\cdec$
and the constants of \eqref{eq:F_bounds} for $\Lyap_p$ on $\mathcal S_p$. For $K_k\to\infty$ with
$\sum_k\gamma_k/K_k<\infty$, the bias vanishes and $\sum_k\gamma_k\|\Delta^{\mathrm{is}}_k\|<\infty$,
which are the properties of the inner-loop error used in Steps~2-4
(Proposition~\ref{prop:tracking}). Assumption~\ref{ass:is} replaces Lemma~\ref{lemma:statspace},
since $\widehat S_K$ does not necessarily lie in $\mathcal S$. Moreover, $\widehat S_K$ is unbounded.
Hence, the bounded Monte Carlo noise of Lemma~\ref{lemma:noise} is replaced by a noise with bounded
conditional second moment.

\section{Numerical Experiments}
\label{sec:experiments}
This section illustrates how the theoretical results for VAE-NLME (Algorithm~\ref{alg:vae_nlme}) and the ELBO VAE-NLME algorithm are reflected in numerical experiments.
\\
A comprehensive empirical evaluation of the VAE-NLME algorithm is provided in a separate benchmark study against SAEM and FOCEI, comprising 1,224 fits across 16 pharmacokinetic and pharmacodynamic models \cite{RohleffBenchmarkingVariationalAutoencoder26}. This benchmark assesses the algorithm's accuracy, robustness, and computational efficiency. The present section therefore focuses exclusively on validating the theoretical results using a simpler example, the real theophylline dataset \cite{PinheiroMixedeffectsModelsSPLUS00}, a standard benchmark in NLME modeling. Its low dimensionality allows key quantities, such as $\vq^\star(\theta)$ and $\nabla_\theta\mathcal L^\star$, to be computed with high accuracy.
\\
First, the model and the convergence of the population parameters are presented. Second, the theorems are validated in an idealized setting and then with the practical implementation used in applications. Finally, the inner-loop error and the gaps introduced in Section~\ref{sec:convergence} are compared.
\\
The VAE-NLME algorithm, with the practical realization of Remark~\ref{rem:practice}, is
available in VAE Studio, the software of the benchmark \cite{RohleffBenchmarkingVariationalAutoencoder26}.
VAE Studio is a desktop application for Windows, macOS and Linux with a graphical user interface, in
which the algorithm can be tried out in a highly optimized implementation. As of November 2026, VAE
Studio is in beta, and a free, time-limited trial license is available on request at
\url{https://www.math.uni-konstanz.de/vae-studio}.
\subsection{The Model}
The theophylline data contain $N=12$ individuals, each with $n_i=10$ concentration measurements after
a single oral dose. The structural model is a one-compartment model with first-order absorption. For a dose $D$, it
reads
\begin{align*}
\frac{dAbs(t)}{dt} &= -k_a\,Abs(t),
&\qquad Abs(0) &= D,\\
\frac{dA(t)}{dt} &= k_a\,Abs(t)-k_e\,A(t),
&\qquad A(0) &= 0,
\end{align*}
with the individual parameters $\phi=(k_a,k_e,V)$ and the concentration $f(t,\phi)=A(t)/V$ as structural model
in \eqref{eq:NLME_model}. The decoder uses the analytical solution
\begin{align}
    \label{eq:theo_C}
f(t,\phi)=\frac{D\,k_a}{V(k_a-k_e)}\left(e^{-k_e t}-e^{-k_a t}\right).
\end{align}
\\
For individual $i$, the dose is $D_i$ and the individual parameters are
$\phi_i=(k_{a,i},k_{e,i},V_i)$. They are positive, i.e., $\Phi=\Reals^3_{>0}$. Hence,
$h=\log$, and each parameter is log-normal, with the body weight as covariate on the volume,
\begin{align*}
  \log(k_{a,i}) &= \log(k_{a,\mathsf{pop}}) + \eta_{a,i},
    &\eta_{a,i}&\sim\mathcal N\big(0,\omega_{k_a}^2\big), \\
  \log(k_{e,i}) &= \log(k_{e,\mathsf{pop}}) + \eta_{e,i},
    &\eta_{e,i}&\sim\mathcal N\big(0,\omega_{k_e}^2\big), \\
  \log(V_{i}) &= \log(V_{\mathsf{pop}}) + \beta_V\,c_i + \eta_{V,i},
    &\eta_{V,i}&\sim\mathcal N\big(0,\omega_{V}^2\big),
    \qquad c_i = \log\!\Big(\frac{W_i}{W_\mathsf{pop}}\Big),
\end{align*}
where $W_i$ is the body weight of individual $i$ and $W_\mathsf{pop}$ the mean body weight. In the
notation of \eqref{eq:individual_parameter_matrix}, $z_i=C_i\zpop+\eta_i$ with
\begin{align*}
C_i = \begin{pmatrix} 1 & 0 & 0 & 0 \\ 0 & 1 & 0 & 0 \\ 0 & 0 & 1 & c_i \end{pmatrix},
\qquad
\Omega = \mathrm{diag}\big(\omega_{k_a}^2, \omega_{k_e}^2, \omega_V^2\big), \\[4pt]
\zpop = \big(\log(k_{a,\mathsf{pop}}), \log(k_{e,\mathsf{pop}}), \log(V_{\mathsf{pop}}), \beta_V\big).
\end{align*}
Together with the residual error parameter $\sigma$, the eight population parameters
$\theta=(\zpop,\Omega,\sigma^2)$ are estimated. The parameters $k_{a,\mathsf{pop}}$, $k_{e,\mathsf{pop}}$
and $V_{\mathsf{pop}}$ are reported on the original scale, i.e., the medians of the log-normal prior, and $\beta_V$
and $\omega$ on the scale of $z$.

\subsection{Convergence of the Population Parameters}
\label{sec:num_pop}
All runs of this section follow Remark~\ref{rem:practice} with adaptively chosen $K_{\mathrm{KL}}$
and $K_{\mathrm{sm}}$ and the settings of Table~\ref{tab:settings}. 
\begin{table}[!ht]
\centering
\small
\renewcommand{\arraystretch}{1.1}
\begin{tabular}{lll}
\toprule
$\nin$ & $8$ & Adam steps per iteration\\
$\alpha$ & $5\cdot10^{-3}$ & learning rate of Adam\\
$M$ & $1$ & samples for the ELBO gradient\\
$K$ & $10$ & samples for the weighted statistics\\
$\nu$ & $5$ & degrees of freedom of the proposal\\
\bottomrule
\end{tabular}
\caption{Settings of all runs on the theophylline data.}
\label{tab:settings}
\end{table}

\begin{figure}[!ht]
    \centering
    \includegraphics[width=\textwidth]{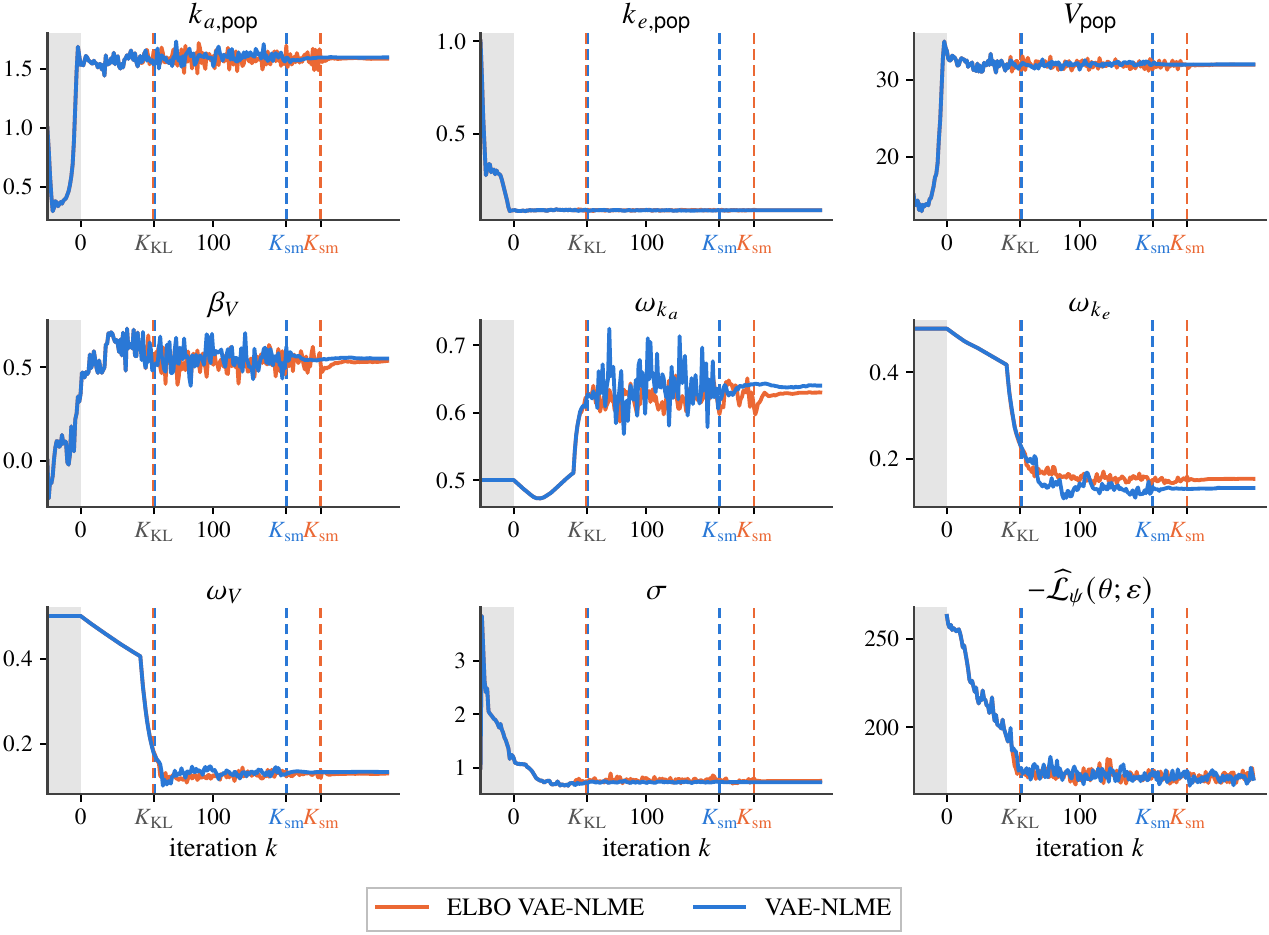}
    \caption{Population parameters $\theta^{(k)}$ on the theophylline data for VAE-NLME (Algorithm~\ref{alg:vae_nlme},
    $-2\mathcal{LL}=\NumStPracLL$) and the ELBO VAE-NLME
    algorithm ($-2\mathcal{LL}=\NumStElboLL$), and the Monte Carlo estimate of the negative
    ELBO in the last panel. The run of VAE-NLME, which stops after
    $\NumStPracN$ iterations, is continued to the same number of iterations. Shaded is the burn-in.}
    \label{fig:convergence_pop}
\end{figure}

\noindent
The runs of this subsection start
from the initial guess
\begin{align*}
  \phi^{\mathsf{init}}=(k_a,k_e,V)=(1,\,1,\,15),\qquad
  \omega^{\mathsf{init}}=(\omega_{k_a},\omega_{k_e},\omega_V)=(0.5,\,0.5,\,0.5),\qquad
  \sigma^{\mathsf{init}}=1 .
\end{align*}
The covariate effect $\beta_V$ needs no initial value and starts at zero.
Figure~\ref{fig:convergence_pop} compares VAE-NLME (Algorithm~\ref{alg:vae_nlme}) with the ELBO VAE-NLME
algorithm. \\
The VAE-NLME algorithm fluctuates more, since
the weighted statistics have a larger variance. The smoothing phase averages out the fluctuations,
and all components of $\theta$ converge. VAE-NLME attains the larger
log-likelihood, as expected from
\eqref{eq:ll_stationarity}.

\FloatBarrier
\subsection{Convergence to a Stationary Point}
\label{sec:num_stationarity}
The two reference points $\theta^\star_{\mathrm{ELBO}}$, stationary
for $\mathcal L^\star$, and $\theta_{\mathrm{ML}}$, stationary for $\mathcal{LL}$, determine the
residual floor of each other by \eqref{eq:ll_stationarity},
\begin{align}
\label{eq:two_floors}
\big|\nabla_\theta\mathcal{LL}(\theta^\star_{\mathrm{ELBO}})\big|
&=\big|\nabla_\theta\mathcal G_{\mathrm{var}}(\theta^\star_{\mathrm{ELBO}})\big|=\NumFloorLL,
\\
\big|\nabla_\theta\mathcal L^\star(\theta_{\mathrm{ML}})\big|
&=\big|\nabla_\theta\mathcal G_{\mathrm{var}}(\theta_{\mathrm{ML}})\big|=\NumFloorLstar.
\end{align}
The two
points agree to three digits, and $\mathcal G_{\mathrm{var}}(\theta^\star_{\mathrm{ELBO}})=\NumGvar$,
i.e., the Gaussian variational family is nearly exact, but its gradient does not vanish.

\subsubsection{Theoretical Validation}
\begin{figure}[!ht]
    \centering
    \includegraphics[width=\textwidth]{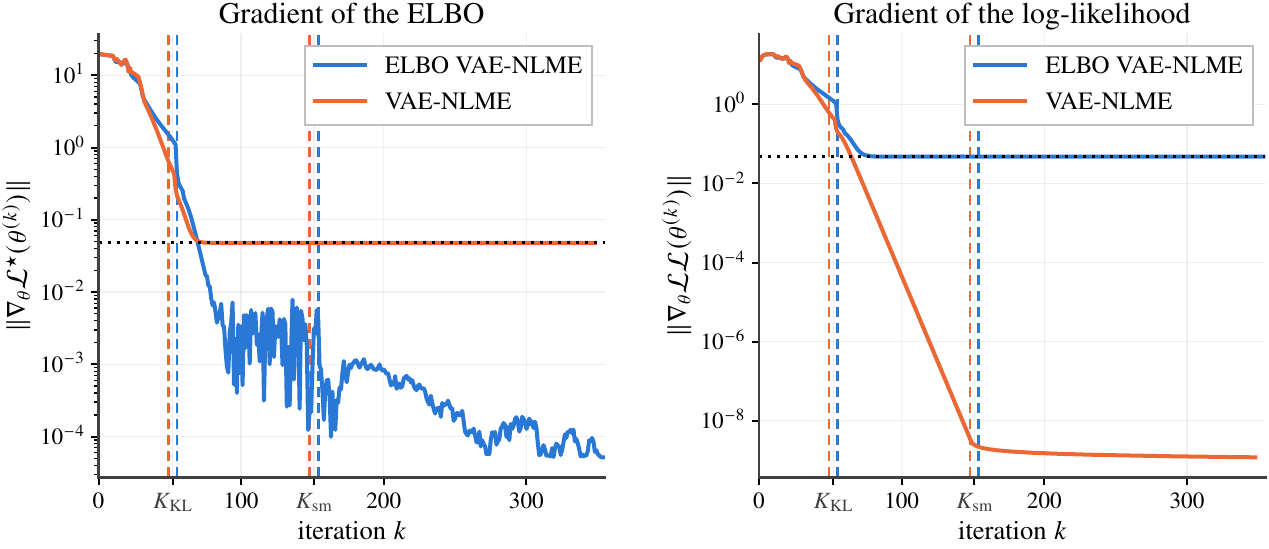}
    \caption{Gradients of the ELBO and the log-likelihood with the exact E-step and $K\to\infty$.
Left: Theorem~\ref{theo:convergence}. Right: Theorem~\ref{theo:is}. Both panels show
the results for both algorithms. The dotted lines indicate the floor
\eqref{eq:two_floors}, corresponding to the gradient of the variational gap.}
    \label{fig:stationarity}
\end{figure}

First, an idealized setting is considered, with the exact E-step, i.e.,
$\vq^{(k)}=\vq^\star(\theta^{(k-1)})$ and inner-loop error $\delta_k=0$. Since there are only $n_z=3$ individual parameters, the maximizer
$\vq^\star(\theta)$ and the residuals
$\|\nabla_\theta\mathcal L^\star(\theta^{(k)})\|$ and
$\|\nabla_\theta\mathcal{LL}(\theta^{(k)})\|$ can be computed to near
machine precision at every iteration using Gauss-Hermite quadrature
and a deterministic optimizer (Appendix~\ref{app:residuals}). In the variational sufficient statistics, $S_3$ uses $M=2^{20}$ fresh samples $\varepsilon$ per iteration, and the weighted statistics are
computed by quadrature, which corresponds to $K\to\infty$.\\
In Figure~\ref{fig:stationarity}, the gradient of $\mathcal L^\star$ tends to zero along the ELBO
VAE-NLME algorithm by Theorem~\ref{theo:convergence}, and the gradient of $\mathcal{LL}$ along
VAE-NLME (Algorithm~\ref{alg:vae_nlme}) by Theorem~\ref{theo:is}. The gradient of the other objective settles at
its floor \eqref{eq:two_floors}.

\begin{figure}[!ht]
    \centering
    \includegraphics[width=\textwidth]{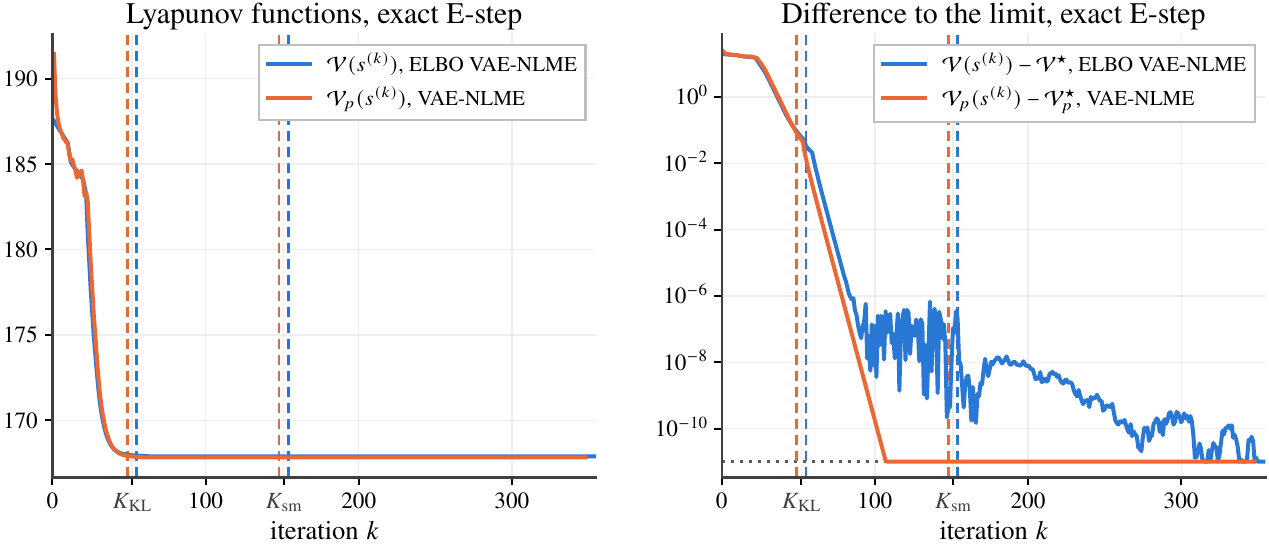}
    \caption{Lyapunov functions of the ELBO VAE-NLME algorithm and of VAE-NLME (Algorithm~\ref{alg:vae_nlme})
    with the exact E-step. Right, the difference to the value $\Lyap^\star$ and $\Lyap_p^\star$ at the
    limit of the run. Differences below the numerical accuracy of $10^{-11}$ are shown on the dotted
    line.}
    \label{fig:lyapunov}
\end{figure}

\noindent
Figure~\ref{fig:lyapunov} shows the Lyapunov functions $\Lyap(s^{(k)})=-\mathcal L^\star(\theta^{(k)})$
of Lemma~\ref{lemma:lyapunov} and $\Lyap_p(s^{(k)})=-\mathcal{LL}(\theta^{(k)})$ of
Section~\ref{sec:is_proof}. Apart from the KL annealing, $\Lyap_p$ decreases monotonically until it
reaches the numerical accuracy. The function $\Lyap$ is not monotone because of the Monte Carlo noise
$\Delta^{\mathrm{mc}}_k$ in \eqref{eq:rm_form}, but its largest increase is only $\NumThLyapVAfterKL$.

\FloatBarrier
\subsubsection{Practical Validation}
\begin{figure}[htb!]
    \centering
    \includegraphics[width=\textwidth]{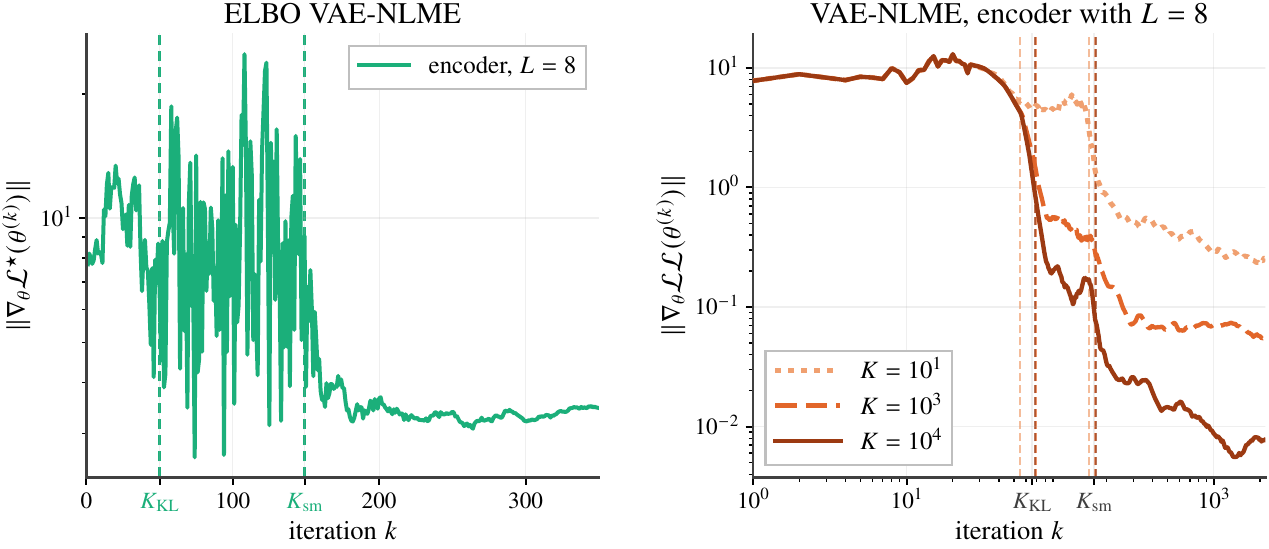}
    \caption{Both algorithms with the encoder. Left: The gradient of $\mathcal L^\star$ for the ELBO
    VAE-NLME algorithm, Theorem~\ref{theo:inexact}. Right: The gradient of the log-likelihood for
    VAE-NLME (Algorithm~\ref{alg:vae_nlme}) with finite $K$, Theorem~\ref{theo:is}~(i).}
    \label{fig:stationarity_finite}
\end{figure}

Figure~\ref{fig:stationarity_finite} shows both algorithms with the encoder, i.e., with
$\delta_k>0$ and finite $K$. For the ELBO VAE-NLME algorithm, the gradient of $\mathcal L^\star$
stagnates. This is the floor \eqref{eq:inexact_floor} of Theorem~\ref{theo:inexact}. For
VAE-NLME (Algorithm~\ref{alg:vae_nlme}), the smoothing phase is extended to $2000$ iterations, since the
remaining noise averages out slowly. The remaining gradient of $\mathcal{LL}$ decreases with $K$, in line with the floor
\eqref{eq:is_floor} of Theorem~\ref{theo:is}, although the encoder does not reach
$\vq^\star(\theta)$.

\FloatBarrier
\subsection{The Encoder and the Inner-Loop Error}
\label{sec:num_inner_loop}
\begin{figure}[!ht]
    \centering
    \includegraphics[width=\textwidth]{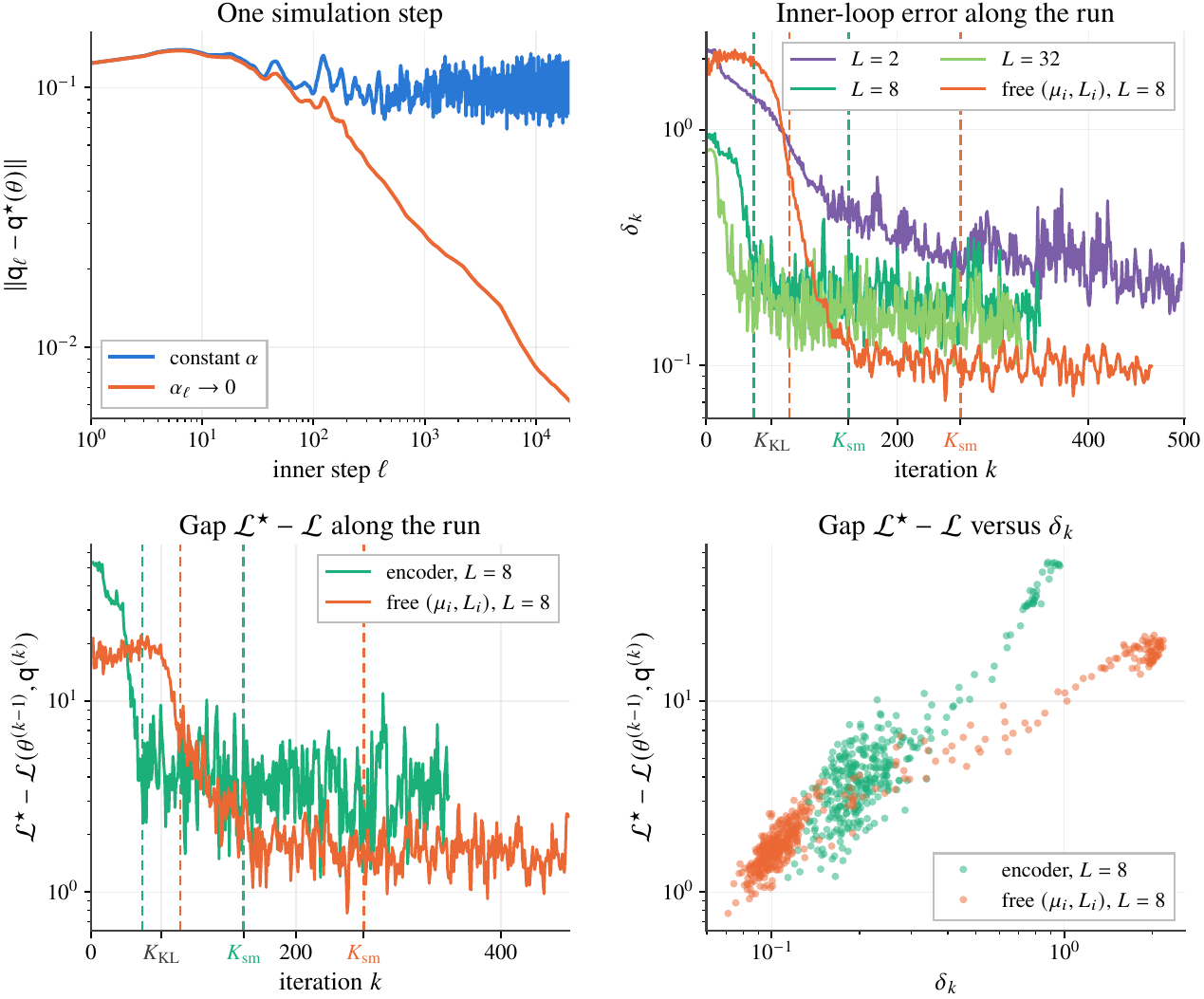}
    \caption{Inner-loop error of the ELBO VAE-NLME algorithm with the encoder and with free
    $(\mu_i,L_i)$; Theorem~\ref{theo:inexact}. (a) One simulation step at fixed $\theta$ on the free
    $(\mu_i,L_i)$, started $\NumEncStartDist$ from $\vq^\star(\theta)$, with fresh samples.
    (b) $\delta_k$ along the run. (c) The gap $\mathcal L^\star-\mathcal L$. (d) The gap against $\delta_k$.}
    \label{fig:inner_loop}
\end{figure}

Figure~\ref{fig:inner_loop} examines the simulation step of the ELBO VAE-NLME algorithm in
Figure~\ref{fig:stationarity_finite}. It compares the encoder with freely optimizing the variational
parameters $\vq_i=(\mu_i,L_i)$ of every individual, as in the non-amortized variational EM algorithm
of \cite{TarekFittingLargeNonlinear26}, with the same number $\nin$ of Adam steps per iteration.
\\
The practical algorithm of Remark~\ref{rem:practice} keeps the step size $\alpha$ of Adam fixed. A
fixed step size leaves a noise ball of radius $O(\alpha)$ around $\vq^\star(\theta)$, so that in
practice the simulation step never reaches $\vq^\star(\theta)$. Hence, $\delta_k$ does not vanish,
even for long inner loops. This explains why the gradient of $\mathcal L^\star$ in
Figure~\ref{fig:stationarity_finite} stagnates at the floor \eqref{eq:inexact_floor} of
Theorem~\ref{theo:inexact}.
\\
The encoder converges faster than the free optimization. It reaches its plateau after about
$\NumEncPlateauEnc$ instead of $\NumEncPlateauFree$ iterations, since every step on the
shared weights moves all individuals at once. In return, its plateau is higher. With its shared
weights, the encoder carries both the amortization gap and the optimization gap of
Section~\ref{sec:amortization}, whereas the free parameters only carry the optimization gap. By
\eqref{eq:gap_delta}, the larger $\delta_k$ of the encoder also means a larger gap
$\mathcal L^\star-\mathcal L=\mathcal G_{\mathrm{am}}+\mathcal G_{\mathrm{opt}}$.

\FloatBarrier
\subsection{ELBO, Likelihood and the Gaps}
\label{sec:num_is}
Table~\ref{tab:elbo_ll} summarizes the gaps of Sections~\ref{sec:is}
and~\ref{sec:amortization} at the limit of the two runs of the ELBO VAE-NLME algorithm in
Figure~\ref{fig:inner_loop} and of one run of VAE-NLME (Algorithm~\ref{alg:vae_nlme}). In all three runs, the
variational gap $\mathcal G_{\mathrm{var}}=\mathcal{LL}-\mathcal L^\star$ is $\NumGvar$.
VAE-NLME has a gap $\mathcal L^\star-\mathcal L$ of the same order as the ELBO VAE-NLME algorithm,
but a gradient of $\mathcal{LL}$ that is smaller by an order of magnitude.

\begin{table}[htb!]
\centering
\small
\setlength{\tabcolsep}{4.5pt}
\begin{tabular}{lrrrrrrr}
\hline
 & $\mathcal L$ & $\mathcal L^\star$ & $\mathcal{LL}$
 & $\mathcal L^\star-\mathcal L$ & $\mathcal G_{\mathrm{var}}$
 & $\|\nabla\mathcal L^\star\|$ & $\|\nabla\mathcal{LL}\|$ \\
\hline
\NumEncTab
\hline
\end{tabular}
\caption{ELBO, marginal log-likelihood, and corresponding gaps at the limit of three runs and at
the two reference points. $\mathcal L=\mathcal L(\theta,\vq)$ is the attained ELBO with the variational parameters $\vq$ of the last iteration,
$\mathcal L^\star=\mathcal L(\theta,\vq^\star(\theta))$ the ELBO at the variational maximizer, and all
gradients are taken with respect to $\theta$.}
\label{tab:elbo_ll}
\end{table}

\section{Conclusion}
\label{sec:conclusion}
The VAE-NLME algorithm converges to the maximum likelihood estimator, although VAEs only maximize
the ELBO. More precisely, with its importance-weighted sufficient statistics,
it converges almost surely to stationary points of the marginal log-likelihood and locally to the
maximum likelihood estimator, up to $O(1/K)$ for $K$ samples (Theorem~\ref{theo:is}), without any
assumption on the training of the encoder. The encoder is trained on the ELBO by stochastic gradients
and serves as proposal. Thus, the ELBO provides the speed of gradient-based training, and the
importance weighting moves the limit from the maximum of the ELBO to the maximum likelihood
estimator. The better the encoder approximates the posterior distribution, the fewer samples are
needed \cite{ChatterjeeSampleSizeRequired18}. The additional
cost is $K$ decoder evaluations per individual and iteration.
\\
Already the ELBO VAE-NLME algorithm converges, although the training of its neural network encoder
is not assumed to converge. The reason is the way the network is embedded into the VAE and the
algorithm. The analysis works with the outputs of the encoder instead of its weights, and the M-step
only uses the sufficient statistics. Hence, an inexact training only perturbs the statistics. This
perturbation vanishes for a contracting inner loop (Theorem~\ref{theo:convergence}) and otherwise
leaves a remaining gradient that is at most proportional to the inner-loop error
(Theorem~\ref{theo:inexact}), independently of the network architecture.
\\
The numerical experiments confirm all three theorems, and a separate benchmark confirms the
accuracy, robustness and speed of the algorithm in practice
\cite{RohleffBenchmarkingVariationalAutoencoder26}. In PMX, VAEs are currently known for inaccurate
solutions. The ELBO VAE-NLME algorithm of the previous work
\cite{RohleffRedefiningParameter25,RohleffVariationalAutoencodersNeural26} is fast, but converges only
to stationary points of the ELBO (Theorem~\ref{theo:convergence}). The importance weighting closes
this gap to the maximum likelihood estimator. Since it only replaces the statistics, the extensions
of the previous work carry over. In particular, the covariate selection of \cite{RohleffRedefiningParameter25} runs exactly as before, but
now with the maximum likelihood estimator instead of a maximum of the ELBO. As the
algorithm is trained by gradients, it also can incorporate neural network components, which can hardly be
trained without gradients due to their many degrees of freedom
\cite{RohleffEvolutionNonlinearMixed26}. This makes the VAE-NLME algorithm a promising method for the
next generation of NLME models in PMX. Beyond PMX, the proof of Theorem~\ref{theo:is} only uses the
exponential family form of the complete-data likelihood. Hence, the algorithm maximizes the
log-likelihood instead of the ELBO also for general marginalized likelihood problems of this form.

\section*{Acknowledgments}
The authors acknowledge the assistance of Claude (Anthropic) in the preparation of this paper, in
particular in discussing ideas, developing proof sketches and preparing the figures of the
numerical experiments. ChatGPT (OpenAI) and Claude were also used to improve the
readability and clarity of the text. The authors take full responsibility for the content of this
paper.

\newpage
\printbibliography
\newpage
\appendix
\section{Appendix}
\subsection{Reduction to the Sufficient Statistics}
\label{app:reduction}

The form \eqref{eq:Q_form} is preserved along the recursion
\eqref{eq:stochastic_approximation}. Assume that at iteration $k-1$ the auxiliary function is of
the form
\begin{align*}
  Q^{(k-1)}(\theta)\;=\;\big\langle s^{(k-1)},\zeta(\theta)\big\rangle-\kappa(\theta)+b^{(k-1)},
\end{align*}
with $s^{(k-1)}\in\Reals^{\dS}$ and a scalar $b^{(k-1)}\in\Reals$ independent of $\theta$. This holds
for $k=1$ if $Q^{(0)}$ is initialized in this form. Inserting the decomposition
\eqref{eq:exponential_family} into \eqref{eq:stochastic_approximation} gives
\begin{align*}
Q^{(k)}(\theta)
&=(1-\gamma_k)\,Q^{(k-1)}(\theta)+\gamma_k\,\widehat{\mathcal L}_{\psi^{(k)}}\big(\theta;\varepsilon^{(k)}\big)\\[2pt]
&=(1-\gamma_k)\Big[\big\langle s^{(k-1)},\zeta(\theta)\big\rangle-\kappa(\theta)+b^{(k-1)}\Big]\\
&\qquad+\gamma_k\Big[\big\langle S(x,\psi^{(k)};\varepsilon^{(k)}),\zeta(\theta)\big\rangle-\kappa(\theta)
 +\mathcal H(\psi^{(k)};\varepsilon^{(k)})+\mathcal C\Big]\\[2pt]
&=\big\langle s^{(k)},\zeta(\theta)\big\rangle-\kappa(\theta)+b^{(k)} ,
\end{align*}
i.e., the form \eqref{eq:Q_form} is preserved, with
\begin{align*}
  s^{(k)} &= s^{(k-1)} + \gamma_k \Big( S\big(x, \psi^{(k)}; \varepsilon^{(k)}\big) - s^{(k-1)} \Big), \\
  b^{(k)} &= b^{(k-1)} + \gamma_k \Big( \mathcal H\big(\psi^{(k)};\varepsilon^{(k)}\big)+\mathcal C - b^{(k-1)} \Big).
\end{align*}

\subsection{Preliminaries: Statistics Space and M-Step Operator}
\label{app:prelim}

\subsubsection*{Proof of Lemma~\ref{lemma:statspace} (The statistics space)}

\begin{proof}
By Assumption~\ref{ass:regularity}(i),
\begin{align*}
  \big|r_{ij}(z)\big|\;\le\;\frac{|x_{ij}|+\sup|f|}{g_{\min}}\;<\;\infty
  \qquad\text{for all } z\in\Reals^{n_z},
\end{align*}
so the third block of $S$ is bounded uniformly in $\varepsilon$, while
the first two are continuous in $\vq$ on the compact set $\mathcal Q$ and free of $\varepsilon$. Hence
\begin{align}
  \label{eq:CS}
  C_S\;:=\;\sup_{\vq\in\mathcal Q,\ \varepsilon}\big\|S(x,\vq;\varepsilon)\big\|\;<\;\infty ,
\end{align}
and $\mathcal S$, the closed convex hull of a bounded set, is compact and convex.
\\
At a single evaluation $S_{1,i}=\mu_i$ and $S_2=\sum_i(\mu_i\mu_i^\top+\Sigma_i)$, so
\begin{align*}
  \Sigtot\big(S(x,\vq;\varepsilon)\big)\;=\;\sum_{i=1}^N\Sigma_i\;\succeq\;N\tau_{\min} I .
\end{align*}
The map $v\mapsto vv^\top$ is convex for the Loewner order, hence $\Sigtot$ is concave for it, and for
every convex combination $s=\sum_m\alpha_m S^{(m)}$ of points in the range of $S$,
\begin{align*}
  \Sigtot(s)\;\succeq\;\sum_m\alpha_m\,\Sigtot\big(S^{(m)}\big)\;\succeq\;N\tau_{\min} I ,
\end{align*}
which extends to the closure by continuity. Likewise $s\mapsto s_3$ is linear and
$S_3\ge s_{\min}$ pointwise by Assumption~\ref{ass:regularity}(iv), so $s_3\ge s_{\min}$ on
$\mathcal S$. Finally $\gamma_k\in[0,1]$ turns \eqref{eq:sa_statistics} into the convex combination
\begin{align*}
  s^{(k)}\;=\;(1-\gamma_k)\,s^{(k-1)}\;+\;\gamma_k\,S\big(x,\vq^{(k)};\varepsilon^{(k)}\big) ,
\end{align*}
so $s^{(k)}\in\mathcal S$ follows by induction from $s^{(0)}\in\mathcal S$.
\end{proof}

\subsubsection*{Proof of Lemma~\ref{lemma:mop} (The M-step operator)}

\begin{proof}
\emph{(i).} $H_s=-\nabla^2_\theta J(\theta_s,s)\succ0$ by Assumption~\ref{ass:mstep}. The surrogate
$J$ is smooth in $\theta$ on $\Theta$ and $\mathcal M$ is continuous, so $s\mapsto H_s$ is continuous
on the compact set $\mathcal S$, and
so are $s\mapsto\|H_s\|$ and, by $H_s\succ0$, $s\mapsto\|H_s^{-1}\|$. Hence
\begin{align*}
  \sup_{s\in\mathcal S}\big\|H_s\big\|\;<\;\infty ,
  \qquad
  \sup_{s\in\mathcal S}\big\|H_s^{-1}\big\|\;<\;\infty .
\end{align*}

\emph{(ii).} $\Theta_{\mathcal S}=\mathcal M(\mathcal S)$ is the continuous image of a compact set,
hence compact, and lies in $\Theta$. The continuous positive functions
$\theta\mapsto\lambda_{\min}(\Omega),\ \|\Omega\|^{-1},\ \sigma^2,\ \sigma^{-2}$ therefore attain
positive minima on it, so for some $c>0$
\begin{align*}
  c\,I\;\preceq\;\Omega\;\preceq\;c^{-1}I,
  \qquad
  c\;\le\;\sigma^2\;\le\;c^{-1},
  \qquad \theta=(\zpop,\Omega,\sigma^2)\in\Theta_{\mathcal S} .
\end{align*}
Being $C^1$ on the compact convex set $\mathcal S$, the map $\mathcal M$ is Lipschitz there.

\emph{(iii).} Since $\nabla_\theta J(\theta,s)=\Dzeta(\theta)^\top s-\nabla_\theta\kappa(\theta)$ is
affine in $s$, it holds $\nabla_s\nabla_\theta J=\Dzeta(\theta_s)^\top$, and differentiating the
first-order condition $\nabla_\theta J(\mathcal M(s),s)=0$ with respect to $s$ gives
\begin{align*}
  \nabla^2_\theta J(\theta_s,s)\,\nabla_s\mathcal M(s)+\Dzeta(\theta_s)^\top
  \;=\;-H_s\,\nabla_s\mathcal M(s)+\Dzeta(\theta_s)^\top\;=\;0 ,
\end{align*}
i.e., $\nabla_s\mathcal M(s)=H_s^{-1}\Dzeta(\theta_s)^\top$.
\end{proof}

\subsection{Step 1: The Variational E-Step}
\label{app:step1}

\subsubsection*{Proof of Lemma~\ref{lemma:tube} (Smoothness and the maximizer on $\mathcal U(\theta)$)}

\begin{proof}
On $\mathcal Q$ the entries of $L_i$ satisfy $\ell_{\min}\le (L_i)_{aa}$ and $|(L_i)_{ab}|\le\ell_{\max}$,
and the derivatives of $f$ and $g$ up to order three are bounded by
Assumption~\ref{ass:regularity}(ii), which also justifies differentiating twice under the expectation in
\eqref{eq:elbo_individual}. Hence $\nabla^2_\vq\mathcal L$ is continuous on the compact set
$\Theta_{\mathcal S}\times\mathcal Q$ and
\begin{align*}
  \Kmax\;:=\;\sup_{\Theta_{\mathcal S}\times\mathcal Q}\big\|\nabla^2_\vq\mathcal L\big\|\;<\;\infty .
\end{align*}
Thus, $\nabla_\vq\mathcal L(\theta,\cdot)$ is $\Kmax$-Lipschitz
on $\mathcal Q$, so $\mathcal L(\theta,\cdot)$ is $\Kmax$-smooth.
The set $\mathcal U(\theta)$ is a ball, hence convex, and $r<\ell_{\min}$ places it inside
$\Reals^{Nn_z}\times\mathcal T_+^N$. By Assumption~\ref{ass:branch},
\begin{align*}
  \nabla^2_\vq\mathcal L(\theta,\vq)\;\preceq\;-\Kmin I
  \quad\text{on }\mathcal U(\theta),
  \qquad
  \nabla_\vq\mathcal L\big(\theta,\vq^\star(\theta)\big)\;=\;0 ,
\end{align*}
so $\vq^\star(\theta)$ is the unique maximizer of the strictly concave function $\mathcal L(\theta,\cdot)$
on $\mathcal U(\theta)$.
\end{proof}

\subsubsection*{Proof of Lemma~\ref{lemma:envelope} (The variational envelope identity)}

\begin{proof}
By Assumption~\ref{ass:branch} the map $\vq^\star$ is $C^1$ on an open set containing the compact set
$\Theta_{\mathcal S}$, so its derivative is bounded there and $\vq^\star$ is Lipschitz, and so is
$\bar S=S\circ\vq^\star$, because $S$ is Lipschitz on $\mathcal Q$. Since $\vq^\star(\theta)\in\mathcal Q$,
\begin{align*}
  \bar S(\theta)\;=\;\Exp_{\varepsilon\sim\mathcal N(0,I)}\big[S\big(x,\vq^\star(\theta);\varepsilon\big)\big]\;\in\;\mathcal S
\end{align*}
as an average of points in the range of $S$ over $\mathcal Q$. The chain rule and the first-order
condition $\nabla_\vq\mathcal L(\theta,\vq^\star(\theta))=0$ give
\begin{align*}
  \nabla_\theta\mathcal L^\star(\theta)
  =&\nabla_\theta\mathcal L\big(\theta,\vq^\star(\theta)\big)
   +\nabla_\theta\vq^\star(\theta)^\top\,\underbrace{\nabla_\vq\mathcal L\big(\theta,\vq^\star(\theta)\big)}_{=\,0}\\
   =&\nabla_\theta\mathcal L\big(\theta,\vq^\star(\theta)\big) \notag \\
  =&\nabla_\theta J\big(\theta,\bar S(\theta)\big) \\
   =&\Dzeta(\theta)^\top\bar S(\theta)-\nabla_\theta\kappa(\theta) ,
\end{align*}
where the second line uses $\nabla_\theta\mathcal L(\theta,\vq)=\nabla_\theta J(\theta,S(\vq))$, since
$\theta$ enters the ELBO only through $J$. The right-hand side is Lipschitz on $\Theta_{\mathcal S}$,
since $J$ is smooth and $\bar S$ is Lipschitz on a compact set.
\end{proof}

\subsection{Step 2: Robbins-Monro Form and Inner-Loop Error}
\label{app:step2}

\subsubsection*{Proof of Lemma~\ref{lemma:noise} (Noise conditions)}

\begin{proof}
Let
$\mathcal G_k=\sigma\big(s^{(0)},\psi^{(0)},\{\varepsilon_\ell\}_{\ell\le \nin}^{(1:k)},\varepsilon^{(1)},\dots,\varepsilon^{(k-1)}\big)$
and $\widetilde{\mathcal F}_k=\sigma\big(\mathcal G_k,\varepsilon^{(k)}\big)$. The iterate
$\vq^{(k)}=\vq(\psi_{\nin})$ is a measurable function of $\psi^{(k-1)}$ and of the inner-loop noise of
iteration $k$, hence $\mathcal G_k$-measurable, while $\varepsilon^{(k)}$ is drawn afresh. Therefore
\begin{align*}
  \Exp\big[S\big(x,\vq^{(k)};\varepsilon^{(k)}\big)\,\big|\,\mathcal G_k\big]
  \;=\;\Exp_{\varepsilon\sim\mathcal N(0,I)}\big[S\big(x,\vq^{(k)};\varepsilon\big)\big]
  \;=\;S\big(\vq^{(k)}\big) ,
  \qquad
  \Exp\big[\Delta^{\mathrm{mc}}_k\,\big|\,\mathcal G_k\big]\;=\;0 .
\end{align*}
Since $\|S(x,\vq;\varepsilon)\|\le C_S$ by \eqref{eq:CS}, it follows that
$\|\Delta^{\mathrm{mc}}_k\|\le 2C_S$, so the partial sums
$\mathcal N_n=\sum_{k\le n}\gamma_k\Delta^{\mathrm{mc}}_k$ form a martingale for
$(\widetilde{\mathcal F}_n)$ with
\begin{align*}
  \sum_{k\ge1}\Exp\Big[\big\|\gamma_k\Delta^{\mathrm{mc}}_k\big\|^2\Big]
  \;\le\;4C_S^2\sum_{k\ge1}\gamma_k^2\;<\;\infty ,
\end{align*}
hence $\mathcal N_n$ is $L^2$-bounded and converges almost surely. Finally $\vq^{(k)}$ and
$\vq^\star(\theta^{(k-1)})$ both lie in $\mathcal Q$, on which $S$ is Lipschitz with constant
$\Lip{S}<\infty$, so
\begin{align*}
  \big\|\Delta^{\mathrm{in}}_k\big\|
  \;=\;\big\|S\big(\vq^{(k)}\big)-S\big(\vq^\star(\theta^{(k-1)})\big)\big\|
  \;\le\;\Lip{S}\,\delta_k .
\end{align*}
\end{proof}

\subsubsection*{Proof of Proposition~\ref{prop:tracking} (Vanishing inner-loop error)}

\begin{proof}
By the update $s^{(k-1)}=s^{(k-2)}+\gamma_{k-1}\big(S(x,\vq^{(k-1)};\varepsilon^{(k-1)})-s^{(k-2)}\big)$
of two points in $\mathcal S$,
\begin{align*}
  \big\|s^{(k-1)}-s^{(k-2)}\big\|\;\le\;\gamma_{k-1}\,\mathrm{diam}(\mathcal S),
  \qquad
  \mathrm{diam}(\mathcal S)=\sup_{s,s'\in\mathcal S}\|s-s'\|<\infty .
\end{align*}
Since $\vq^\star$ and $\mathcal M$ are Lipschitz (Lemmas~\ref{lemma:mop} and~\ref{lemma:envelope}),
it follows with $c_1=\Lip{\vq^\star}\Lip{\mathcal M}\,\mathrm{diam}(\mathcal S)$ that
\begin{align*}
\big\|\vq^{(k-1)}-\vq^\star(\theta^{(k-1)})\big\|
&\le \delta_{k-1}+\big\|\vq^\star(\theta^{(k-2)})-\vq^\star(\theta^{(k-1)})\big\| \notag \\
&\le \delta_{k-1}+\Lip{\vq^\star}\Lip{\mathcal M}\big\|s^{(k-1)}-s^{(k-2)}\big\| \\
&\le\delta_{k-1}+c_1\,\gamma_{k-1} .
\end{align*}
Inserting this into the contraction
$\delta_k\le\rho\,\|\vq^{(k-1)}-\vq^\star(\theta^{(k-1)})\|+c_0\alpha_k$ and using
$\alpha_k\le c_\alpha\gamma_k$ (Assumption~\ref{ass:contraction}),
\begin{align*}
  \delta_k\;\le\;\rho\,\delta_{k-1}+\rho\,c_1\gamma_{k-1}+c_0c_\alpha\gamma_k ,
\end{align*}
and with $u_k=\delta_k/\gamma_k$,
\begin{align*}
  u_k\;\le\;\rho\,\frac{\gamma_{k-1}}{\gamma_k}\,u_{k-1}
  +\rho\,c_1\frac{\gamma_{k-1}}{\gamma_k}+c_0c_\alpha .
\end{align*}
Let $\epsilon>0$ with $\rho(1+\epsilon)<1$. Since $\gamma_{k-1}/\gamma_k\to1$ there is $k_0$
with $\gamma_{k-1}/\gamma_k\le1+\epsilon$ for $k\ge k_0$, so
\begin{align*}
  u_k\;\le\;\rho(1+\epsilon)\,u_{k-1}+\big(\rho\,c_1(1+\epsilon)+c_0c_\alpha\big)
  \qquad (k\ge k_0),
\end{align*}
and iterating this affine contraction gives
\begin{align*}
  \limsup_{k\to\infty}\ \frac{\delta_k}{\gamma_k}
  \;\le\;\frac{\rho\,c_1(1+\epsilon)+c_0c_\alpha}{1-\rho(1+\epsilon)} .
\end{align*}
The left-hand side does not depend on $\epsilon$, so letting $\epsilon\downarrow0$ yields
\begin{align*}
  \limsup_{k\to\infty}\ \frac{\delta_k}{\gamma_k}
  \;\le\;\frac{\rho\,c_1+c_0c_\alpha}{1-\rho}\;<\;\infty .
\end{align*}
Thus $\delta_k=O(\gamma_k)\to0$ and
$\sum_k\gamma_k\delta_k=O\big(\sum_k\gamma_k^2\big)<\infty$ by Assumption~\ref{ass:stepsizes}.
\end{proof}

\subsubsection*{Proof of Proposition~\ref{prop:apt} (Asymptotic pseudo-trajectory)}

\begin{proof}
Write the update \eqref{eq:rm_form} as $s^{(k)}=s^{(k-1)}+\gamma_kg_k$ with
$g_k:=\Fmf(s^{(k-1)})+\Delta^{\mathrm{mc}}_k+\Delta^{\mathrm{in}}_k$. By Lemma~\ref{lemma:noise} and
$\vq^{(k)},\vq^\star(\theta^{(k-1)})\in\mathcal Q$,
\begin{align}
  \label{eq:cg}
  \|g_k\|\;\le\;c_g\;:=\;\sup_{\mathcal S}\|\Fmf\|+2C_S+\Lip{S}\,\mathrm{diam}(\mathcal Q) .
\end{align}
Fix $T>0$ and $n\in\mathbb N$, let $m:=\max\{k:t_k\le t_n+T\}$ and put, for $n\le k\le m$,
\begin{align*}
E_k:=\big\|s^{(k)}-\Flow_{t_k-t_n}\big(s^{(n)}\big)\big\| ,
\qquad
W_{n,k}:=\sum_{j=n+1}^{k}\gamma_j\big(\Delta^{\mathrm{mc}}_j+\Delta^{\mathrm{in}}_j\big).
\end{align*}
Summing the update over $j=n+1,\dots,k$ gives
$s^{(k)}=s^{(n)}+\sum_{j=n+1}^{k}\gamma_j\Fmf(s^{(j-1)})+W_{n,k}$. Subtracting the integral form
$\Flow_{t_k-t_n}(s^{(n)})=s^{(n)}+\int_{t_n}^{t_k}\Fmf(\Flow_{u-t_n}(s^{(n)}))\,du$, and using
$\|\Flow_{t_{j-1}-t_n}(s^{(n)})-\Flow_{u-t_n}(s^{(n)})\|\le c_{\mathcal F}\gamma_j$ for $u\in[t_{j-1},t_j]$ with
$c_{\mathcal F}=\sup_{\mathcal S}\|\Fmf\|$,
\begin{align*}
E_k
&\le\sum_{j=n+1}^{k}\int_{t_{j-1}}^{t_j}\Big\|\Fmf\big(s^{(j-1)}\big)-\Fmf\big(\Flow_{u-t_n}(s^{(n)})\big)\Big\|\,du
 \;+\;\big\|W_{n,k}\big\| \notag \\
&\le\Lip{\Fmf}\sum_{j=n+1}^{k}\gamma_j E_{j-1}
 \;+\;\Lip{\Fmf}c_{\mathcal F}\sum_{j=n+1}^{k}\gamma_j^2
 \;+\;\sup_{n\le l\le k}\big\|W_{n,l}\big\| .
\end{align*}
With $E_n=0$ and $\sum_{j=n+1}^m\gamma_j\le T+\gamma_{n+1}$, the discrete Gronwall inequality gives
\begin{align}
  \label{eq:gronwall}
  \max_{n\le k\le m} E_k
  \;\le\;
  e^{\Lip{\Fmf}(T+\gamma_{n+1})}\Big(\Lip{\Fmf}c_{\mathcal F}\sum_{j>n}\gamma_j^2+\sup_{n\le l\le m}\big\|W_{n,l}\big\|\Big),
\end{align}
where
\begin{align*}
\sup_{n\le l\le m}\big\|W_{n,l}\big\|
\;\le\;\underbrace{\sup_{n\le l\le m}\Big\|\sum_{j=n+1}^{l}\gamma_j\Delta^{\mathrm{mc}}_j\Big\|}_{\to\,0\ \text{a.s.\ (Lemma~\ref{lemma:noise})}}
\;+\;\underbrace{\Lip{S}\sum_{j>n}\gamma_j\delta_j}_{\to\,0\ \text{(Proposition~\ref{prop:tracking})}} .
\end{align*}
Since also $\sum_j\gamma_j^2<\infty$, the right-hand side of \eqref{eq:gronwall} tends to $0$ almost
surely, which is \eqref{eq:apt} at the grid points. For $t\in[t_n,t_{n+1}]$ the interpolation differs
from $s^{(n)}$ by at most $\gamma_{n+1}\|g_{n+1}\|\le c_g\gamma_{n+1}\to0$, and $\Flow$ is Lipschitz
in time, so the conclusion holds uniformly. The limit set of a bounded asymptotic pseudo-trajectory of
a Lipschitz semiflow is non-empty, compact, connected and invariant
\cite{BorkarStochasticApproximationDynamical23,StochasticApproximationRecursive03}, and boundedness
holds because $\mathcal S$ is compact.
\end{proof}

\subsection{Step 3: Lyapunov Stability}
\label{app:step3}

\subsubsection*{Proof of Lemma~\ref{lemma:lyapunov} (Lyapunov decay)}
\begin{proof}
Three identities are combined. First, $\mathcal M(s)$ is a stationary point of $J(\cdot,s)$, so
\eqref{eq:m5} reads
\begin{align}
\label{eq:mstep_foc}
\Dzeta(\theta_s)^\top s \;=\; \nabla_\theta \kappa(\theta_s).
\end{align}
Second, evaluating the envelope identity \eqref{eq:envelope} at $\theta_s$ and inserting
\eqref{eq:mstep_foc} expresses the gradient of $\mathcal L^\star$ through the mean field,
\begin{align}
\label{eq:var_fisher}
\begin{aligned}
\nabla_\theta \mathcal{L}^\star(\theta_s)
&= \Dzeta(\theta_s)^\top \bar{S}(\theta_s) - \nabla_\theta\kappa(\theta_s) \\
&= \Dzeta(\theta_s)^\top\big(\bar{S}(\theta_s) - s\big) \\
&= \Dzeta(\theta_s)^\top \Fmf(s).
\end{aligned}
\end{align}
Third, the sensitivity \eqref{eq:mstep_sens} and the chain rule give
\begin{align}
\label{eq:grad_V}
\nabla_s\Lyap(s)
\;=\;-\,\nabla_s\mathcal M(s)^\top\nabla_\theta\mathcal L^\star(\theta_s)
\;=\;-\,\Dzeta(\theta_s)H_s^{-1}\nabla_\theta\mathcal L^\star(\theta_s).
\end{align}
Since $J$ is smooth in $(\theta,s)$, $\mathcal M$ is Lipschitz on $\mathcal S$ and
$\sup_{s}\|H_s^{-1}\|<\infty$, each factor in \eqref{eq:grad_V} is Lipschitz on the compact set
$\mathcal S$, and so is $\nabla_s\Lyap$. Combining \eqref{eq:var_fisher} and \eqref{eq:grad_V},
\begin{align*}
\big\langle \nabla_s\Lyap(s),\,\Fmf(s)\big\rangle
&=-\,\nabla_\theta \mathcal{L}^\star(\theta_s)^\top H_s^{-1}\,\Dzeta(\theta_s)^\top \Fmf(s) \\
&=-\,\nabla_\theta \mathcal{L}^\star(\theta_s)^\top H_s^{-1}\,\nabla_\theta \mathcal{L}^\star(\theta_s) \\
&=-\big\|\nabla_\theta \mathcal{L}^\star(\theta_s)\big\|_{H_s^{-1}}^2 ,
\end{align*}
which is \eqref{eq:decay}, since $H_s^{-1}\succ0$. The bounds \eqref{eq:F_bounds} hold with
$\cdec=\inf_{s\in\mathcal S}\|H_s\|^{-1}>0$ and
$\cgrad=\sup_{s\in\mathcal S}\|\Dzeta(\theta_s)H_s^{-1}\|<\infty$, both finite and positive by
Lemma~\ref{lemma:mop} and compactness of $\mathcal S$.
\end{proof}

\subsubsection*{Proof of Proposition~\ref{prop:discrete_decay} (Discrete Lyapunov decay)}

\begin{proof}
As in the proof of Proposition~\ref{prop:apt}, $s^{(k)}=s^{(k-1)}+\gamma_kg_k$ with
$g_k=\Fmf(s^{(k-1)})+\Delta^{\mathrm{mc}}_k+\Delta^{\mathrm{in}}_k$ and $\|g_k\|\le c_g$. Put
$\cLyap=\sup_{\mathcal S}\|\nabla\Lyap\|<\infty$. Since $\nabla\Lyap$ is Lipschitz on the convex
set $\mathcal S$, the descent lemma, the decay
$\langle\nabla\Lyap(s),\Fmf(s)\rangle\le-\cdec\|\nabla_\theta\mathcal L^\star(\theta_s)\|^2$ of
Lemma~\ref{lemma:lyapunov} and $\|\Delta^{\mathrm{in}}_k\|\le\Lip{S}\delta_k$ give, with
$c_2=\Lip{\nabla\Lyap}c_g^2/2$,
\begin{align}
\Lyap\big(s^{(k)}\big)
&\le \Lyap\big(s^{(k-1)}\big)+\gamma_k\big\langle\nabla\Lyap(s^{(k-1)}),g_k\big\rangle
   +\tfrac{\Lip{\nabla\Lyap}}{2}\gamma_k^2\|g_k\|^2 \notag \\
&\le \Lyap\big(s^{(k-1)}\big)
   -\gamma_k\,\cdec\big\|\nabla_\theta\mathcal L^\star(\theta^{(k-1)})\big\|^2
   +\gamma_k\big\langle\nabla\Lyap(s^{(k-1)}),\Delta^{\mathrm{mc}}_k\big\rangle \notag \\
&\qquad
   +\gamma_k\,\big\|\nabla\Lyap(s^{(k-1)})\big\|\, \Lip{S}\delta_k
   +c_2\gamma_k^2 .
   \label{eq:per_step}
\end{align}
This per-step inequality only uses Assumptions~\ref{ass:regularity}-\ref{ass:branch}. Here,
$\|\nabla\Lyap(s^{(k-1)})\|$ is bounded by $\cLyap$. Summing from $1$ to $n$,
\begin{align}
  \label{eq:summed}
  \Lyap\big(s^{(n)}\big)+\cdec\sum_{k=1}^n\gamma_k\big\|\nabla_\theta\mathcal L^\star(\theta^{(k-1)})\big\|^2
  \;\le\;&
  \Lyap\big(s^{(0)}\big)
  + \underbrace{\sum_{k=1}^n\gamma_k\big\langle\nabla\Lyap(s^{(k-1)}),\Delta^{\mathrm{mc}}_k\big\rangle}_{=:\,\mathcal Z_n}
  \notag \\
  &+ \cLyap\Lip{S}\underbrace{\sum_{k=1}^n\gamma_k\delta_k}_{<\,\infty}
  + c_2\underbrace{\sum_{k=1}^n\gamma_k^2}_{<\,\infty} .
\end{align}
The process $(\mathcal Z_n)$ is a martingale, because $\nabla\Lyap(s^{(k-1)})$ is
$\mathcal G_k$-measurable and $\Exp[\Delta^{\mathrm{mc}}_k|\mathcal G_k]=0$, and
\begin{align*}
  \sum_{k\ge1}\Exp\Big[\big|\mathcal Z_k-\mathcal Z_{k-1}\big|^2\Big]
  \;\le\;4\cLyap^2C_S^2\sum_{k\ge1}\gamma_k^2\;<\;\infty ,
\end{align*}
so $\mathcal Z_n$ converges almost surely. As $\Lyap$ is bounded on $\mathcal S$, the left-hand side of
\eqref{eq:summed} is bounded above, which gives \eqref{eq:V_converges}. Moreover $\Lyap(s^{(n)})$ is
the sum of a convergent martingale, a convergent remainder and a non-increasing term bounded below, and
therefore converges.
\end{proof}

\subsubsection*{Proof of Proposition~\ref{prop:F_to_zero} (Vanishing stationarity residual)}

\begin{proof}
Write $R_k:=\|\nabla_\theta\mathcal L^\star(\theta^{(k)})\|^2$. The map
$s\mapsto\|\nabla_\theta\mathcal L^\star(\mathcal M(s))\|^2$ is Lipschitz on the compact set
$\mathcal S$ with some constant $\Lip{R}$, and $\|s^{(k)}-s^{(k-1)}\|\le c_g\gamma_k$ by
\eqref{eq:cg}, so
\begin{align}
  \label{eq:F_increments}
  \big|R_k-R_{k-1}\big|\;\le\;\Lip{R}c_g\,\gamma_k .
\end{align}
Write $c:=\Lip{R}c_g$, which can be assumed positive, since otherwise $R$ is constant and, by
\eqref{eq:V_converges} together with $\sum_k\gamma_k=\infty$, identically zero.
\\
Let $\epsilon>0$. By Proposition~\ref{prop:discrete_decay} the series $\sum_k\gamma_kR_{k-1}$
converges, so there is $n_1$ with
\begin{align}
  \label{eq:F_tail}
  \sum_{k>n}\gamma_k R_{k-1}\;<\;\frac{\epsilon^2}{c}\qquad\text{for every }n\ge n_1 .
\end{align}
It is shown that $R_n<2\epsilon$ for every $n\ge n_1$. Suppose instead $R_{n_0}\ge2\epsilon$ for some
$n_0\ge n_1$. If $R_k\ge\epsilon$ for all $k\ge n_0$, then
$\sum_{k>n_0}\gamma_kR_{k-1}\ge\epsilon\sum_{k>n_0}\gamma_k=\infty$ by
Assumption~\ref{ass:stepsizes}, contradicting \eqref{eq:F_tail}. Let $m>n_0$ be the smallest index
with $R_m<\epsilon$, so that $R_k\ge\epsilon$ for $n_0\le k<m$. By \eqref{eq:F_increments},
\begin{align*}
\epsilon&\le\;R_{n_0}-R_m\;\le\;\sum_{k=n_0+1}^{m}\big|R_k-R_{k-1}\big|
\;\le\;c\sum_{k=n_0+1}^{m}\gamma_k ,
\end{align*}
hence $\sum_{k=n_0+1}^{m}\gamma_k\ge\epsilon/c$. Since $R_{k-1}\ge\epsilon$ for $n_0<k\le m$,
\begin{align*}
\sum_{k>n_0}\gamma_k R_{k-1}
\;\ge\;\sum_{k=n_0+1}^{m}\gamma_k R_{k-1}
\;\ge\;\epsilon\sum_{k=n_0+1}^{m}\gamma_k
\;\ge\;\frac{\epsilon^2}{c} ,
\end{align*}
again contradicting \eqref{eq:F_tail}. Hence $0\le R_n<2\epsilon$ for all $n\ge n_1$, and since
$\epsilon>0$ was arbitrary, $R_k\to0$ almost surely.
\end{proof}

\subsection{Step 4: Stationary Optimality Conditions}
\label{app:step4}

\subsubsection*{Proof of Lemma~\ref{lemma:limitset} (Characterization of the limit set)}
\begin{proof}
The second identity is the first-order condition of Assumption~\ref{ass:branch}. By the envelope
identity \eqref{eq:envelope},
\begin{align*}
  0\;=\;\nabla_\theta\mathcal L^\star(\theta^\star)
  \;=\;\nabla_\theta\mathcal L(\theta^\star,\vq^\star)
  \;=\;\nabla_\theta J\big(\theta^\star,\bar S(\theta^\star)\big),
\end{align*}
which gives the first and, with $s^\star=\bar S(\theta^\star)$, the third identity. The converse follows
by reading the same chain backwards.
\end{proof}

\subsection{Proofs of Theorems~\ref{theo:convergence}-\ref{theo:is}}
\label{app:main}

\noindent
The main results use two further lemmas. The first one considers a sequence that contracts towards a
level at rate $\gamma_k$, and shows that its limit superior lies below this level, even if the
contraction is perturbed.

\begin{lemma}[Perturbed contraction]
\label{lemma:contraction_rec}
Let $a>0$, $\gamma_k\ge0$ with $\sum_k\gamma_k=\infty$, and $\beta_k\in\Reals$ with
$\limsup_k\beta_k\le\beta$. If $u_k\in\Reals$ satisfies, for all sufficiently large $k$,
$a\gamma_k\le1$ and
\begin{align}
  \label{eq:contraction_rec}
  u_k\;\le\;(1-a\gamma_k)\,u_{k-1}+\gamma_k\,\beta_k ,
\end{align}
then $\limsup_k u_k\le\beta/a$.
\end{lemma}

\begin{proof}
Let $\epsilon>0$, $c=(\beta+\epsilon)/a$, and $k_0$ such that $a\gamma_k\le1$, $\beta_k\le ac$ and
\eqref{eq:contraction_rec} hold for all $k>k_0$. Then
\begin{align*}
  u_k-c
  \;\le\;(1-a\gamma_k)\,u_{k-1}+a\gamma_k\,c-c
  \;=\;(1-a\gamma_k)\,(u_{k-1}-c) ,
\end{align*}
and with $1-x\le e^{-x}$ and $\sum_k\gamma_k=\infty$, the excess $w_k=\max\{u_k-c,0\}$ satisfies
\begin{align*}
  w_k
  \;\le\;(1-a\gamma_k)\,w_{k-1}
  \;\le\;e^{-a\gamma_k}\,w_{k-1}
  \;\le\;w_{k_0}\exp\Big(-a\sum_{j=k_0+1}^{k}\gamma_j\Big)
  \;\xrightarrow[k\to\infty]{}\;0 .
\end{align*}
Hence $\limsup_ku_k\le(\beta+\epsilon)/a$ for every $\epsilon>0$.
\end{proof}

\noindent
The second lemma is the core of the local statements in Theorems~\ref{theo:inexact}
and~\ref{theo:is}. It is formulated for a general objective $f$, which is $\mathcal L^\star$ in
Theorem~\ref{theo:inexact} and $\mathcal{LL}$ in Theorem~\ref{theo:is}. Along the iterates, let
\begin{align*}
  e_k\;=\;f(\theta^\star)-f\big(\theta^{(k)}\big),
  \qquad
  G_k\;=\;\big\|\nabla_\theta f\big(\theta^{(k)}\big)\big\|
\end{align*}
for the gap to the maximum and the norm of the gradient.

\begin{lemma}[Local capture]
\label{lemma:capture}
Let $f$ be continuously differentiable on an open set containing $\Theta'\subset\Theta$, and twice
continuously differentiable near a point $\theta^\star\in\Theta'$ with $\nabla_\theta f(\theta^\star)=0$
and $m=\lambda_{\min}\big(-\nabla^2_\theta f(\theta^\star)\big)>0$. Let $\theta^{(k)}\in\Theta'$ with
$\|\theta^{(k)}-\theta^{(k-1)}\|\to0$ satisfy the descent inequality
\begin{align}
  \label{eq:capture_descent}
  e_k\;\le\;e_{k-1}-\gamma_k\,\cdec\,G_{k-1}^2+\gamma_k\,\cgrad\,G_{k-1}\,b_k+\gamma_k\xi_k
  +c_2\gamma_k^2 ,
\end{align}
where $\gamma_k$ satisfies Assumption~\ref{ass:stepsizes}, $\cdec>0$ and $\cgrad,c_2\ge0$ are
constants, the bias satisfies $b_k\ge0$ and $\limsup_kb_k\le b$, and the noise $\xi_k$ is such that
$\sum_k\gamma_k\xi_k$ converges. Then there is a neighborhood $\mathcal U$ of $\theta^\star$ such
that, if $C_b:=2\cgrad b/(\cdec m)$ is sufficiently small and $\theta^{(k)}\in\mathcal U$ for
infinitely many $k$,
\begin{align*}
  \limsup_{k\to\infty}\ \big\|\theta^{(k)}-\theta^\star\big\|\;\le\;C_b .
\end{align*}
\end{lemma}

\begin{proof}
Write $\Delta_k=\|\theta^{(k)}-\theta^\star\|$.

\emph{Step 1, $f$ near $\theta^\star$.} By continuity of the Hessian, there is $r_\star>0$ with
$\nabla^2_\theta f\preceq-\frac m2I$ on the closed ball of radius $r_\star$ around $\theta^\star$. Let
$\mathcal B=\{\theta\in\Theta':\|\theta-\theta^\star\|\le r_\star\}$, $\theta\in\mathcal B$ and
$t=\|\theta-\theta^\star\|$. Taylor's theorem around $\theta^\star$ and around $\theta$ gives
\begin{align}
  \frac m4\,t^2
  \;\le\;f(\theta^\star)-f(\theta)
  &\;\le\;\big\langle\nabla_\theta f(\theta),\,\theta^\star-\theta\big\rangle-\frac m4\,t^2
  \notag\\
  &\;\le\;\big\|\nabla_\theta f(\theta)\big\|\,t-\frac m4\,t^2
  \;\le\;\frac1m\,\big\|\nabla_\theta f(\theta)\big\|^2 ,
  \label{eq:growth_pl}
\end{align}
where the last step maximizes over $t$. The outer bounds are the quadratic growth and the
Polyak-{\L}ojasiewicz inequality \cite{KarimiLinearConvergenceGradient16}.

\emph{Step 2, contraction in $\mathcal B$.} Let $\theta^{(k-1)}\in\mathcal B$ and $a=\cdec m/2$.
Young's inequality and $G_{k-1}^2\ge m\,e_{k-1}$ from \eqref{eq:growth_pl} give
\begin{align*}
  -\cdec G_{k-1}^2+\cgrad G_{k-1}b_k
  \;\le\;-\frac{\cdec}{2}\,G_{k-1}^2+\frac{(\cgrad b_k)^2}{2\cdec}
  \;\le\;-a\,e_{k-1}+\frac{(\cgrad b_k)^2}{2\cdec} .
\end{align*}
The remainder $\varrho_k=\sum_{j>k}(\gamma_j\xi_j+c_2\gamma_j^2)$ tends to zero and satisfies
$\gamma_k\xi_k+c_2\gamma_k^2=\varrho_{k-1}-\varrho_k$. Inserting both into
\eqref{eq:capture_descent} gives for $u_k=e_k+\varrho_k$
\begin{align}
  u_k
  \;\le\;(1-a\gamma_k)\,e_{k-1}+\gamma_k\frac{(\cgrad b_k)^2}{2\cdec}+\varrho_{k-1}
  \;=\;(1-a\gamma_k)\,u_{k-1}+\gamma_k\beta_k,
  \label{eq:u_rec}
\end{align}
with $\beta_k=\frac{(\cgrad b_k)^2}{2\cdec}+a\varrho_{k-1}$, and, with $v_\infty=\frac m4C_b^2$,
\begin{align*}
  \limsup_{k\to\infty}\beta_k
  \;\le\;\frac{(\cgrad b)^2}{2\cdec}
  \;=\;\frac{\cdec m}{2}\cdot\frac m4\Big(\frac{2\cgrad b}{\cdec m}\Big)^2
  \;=\;a\,v_\infty .
\end{align*}

\emph{Step 3, the iterates stay in $\mathcal B$.} Let $v=\frac m4\big(\frac{r_\star}{2}\big)^2$ and
\begin{align*}
  \mathcal U\;=\;\Big\{\theta\in\Theta':\ \|\theta-\theta^\star\|<\frac{r_\star}{2}\ \text{ and }\
  f(\theta^\star)-f(\theta)<\frac v2\Big\},
\end{align*}
a neighborhood of $\theta^\star$ by continuity of $f$. The constant $C_b$ is called sufficiently small if
$C_b<r_\star/4$, i.e., $v_\infty<v/4$. Since $\gamma_k$, $\varrho_k$ and the steps tend to zero and
$\limsup_k\beta_k<\frac34av$, there is $k_1$ such that, for all $k\ge k_1$,
\begin{align*}
  \big\|\theta^{(k)}-\theta^{(k-1)}\big\|<\frac{r_\star}{2},
  \qquad
  a\gamma_k\le1,
  \qquad
  |\varrho_k|\le\frac v8,
  \qquad
  \beta_k\le\frac34\,a\,v .
\end{align*}
Take $k_2\ge k_1$ with $\theta^{(k_2)}\in\mathcal U$. Then $\Delta_{k_2}<r_\star/2$ and
$u_{k_2}\le\frac v2+\frac v8\le\frac34v$. If $\Delta_{k-1}<r_\star/2$ and $u_{k-1}\le\frac34v$ for
some $k>k_2$, then \eqref{eq:u_rec} and \eqref{eq:growth_pl} give
\begin{align*}
  \Delta_k
  &\;\le\;\Delta_{k-1}+\big\|\theta^{(k)}-\theta^{(k-1)}\big\|
  \;<\;\frac{r_\star}{2}+\frac{r_\star}{2}\;=\;r_\star ,\\
  u_k
  &\;\le\;(1-a\gamma_k)\,\frac34v+\gamma_k\,\frac34av\;=\;\frac34v ,\\
  \frac m4\,\Delta_k^2
  &\;\le\;e_k\;=\;u_k-\varrho_k\;\le\;\frac34v+\frac v8\;<\;v\;=\;\frac m4\Big(\frac{r_\star}{2}\Big)^2 .
\end{align*}
By induction, $\Delta_k<r_\star/2$ and \eqref{eq:u_rec} hold for all $k>k_2$.

\emph{Step 4, the limit.} Lemma~\ref{lemma:contraction_rec} applied to \eqref{eq:u_rec},
$\varrho_k\to0$ and \eqref{eq:growth_pl} give
\begin{align*}
  \frac m4\,\limsup_{k\to\infty}\Delta_k^2
  \;\le\;\limsup_{k\to\infty}e_k
  \;=\;\limsup_{k\to\infty}u_k
  \;\le\;v_\infty
  \;=\;\frac m4\,C_b^2 .
\end{align*}
\end{proof}

\subsubsection*{Proof of Theorem~\ref{theo:convergence} (Convergence of the ELBO VAE-NLME algorithm)}
\begin{proof}
The first limit is Proposition~\ref{prop:F_to_zero}. The set $\Theta_{\mathcal S}$ is compact by
Lemma~\ref{lemma:mop}(ii) and $\theta^{(k)}=\mathcal M(s^{(k)})$ lies in it, so every subsequence of
$\{\theta^{(k)}\}$ has a convergent sub-subsequence $\theta^{(k_j)}\to\hat\theta$, whose limit
satisfies, by continuity of $\nabla_\theta\mathcal L^\star$,
\begin{align*}
  \big\|\nabla_\theta\mathcal L^\star(\hat\theta)\big\|
  \;=\;\lim_{j\to\infty}\big\|\nabla_\theta\mathcal L^\star\big(\theta^{(k_j)}\big)\big\|
  \;=\;0,
  \qquad\text{i.e.,}\qquad
  \hat\theta\in\Theta^\star .
\end{align*}
Hence $\Theta^\star$ is non-empty, and it is compact as a closed subset of the compact set
$\Theta_{\mathcal S}$, since $\nabla_\theta\mathcal L^\star$ is continuous (Lemma~\ref{lemma:envelope}).
This gives the second limit and, by Lemma~\ref{lemma:limitset}, $(\theta^\star,\vq^\star(\theta^\star))$
is a stationary point of the ELBO. For the last statement,
\begin{align*}
  \big\|\theta^{(k)}-\theta^{(k-1)}\big\|
  \;\le\;\Lip{\mathcal M}\,\big\|s^{(k)}-s^{(k-1)}\big\|
  \;\le\;\Lip{\mathcal M}\,c_g\,\gamma_k
  \;\xrightarrow[k\to\infty]{}\;0
\end{align*}
by Lemma~\ref{lemma:mop}(ii) and \eqref{eq:cg}, so the set of limit points of $\{\theta^{(k)}\}$ is
connected \cite{StochasticApproximationRecursive03} and contained in $\Theta^\star$. It reduces to a
single point $\theta^\star$, since $\Theta^\star$ is compact with isolated elements and hence finite.
Finally, the set $L$ of limit points of $\{s^{(k)}\}$ is compact and invariant by
Proposition~\ref{prop:apt}, and $\mathcal M(L)=\{\theta^\star\}$ by continuity of $\mathcal M$. On
$L$, the ODE $\dot s=\bar S(\mathcal M(s))-s$ therefore reads $\dot s=s^\star-s$ with
$s^\star=\bar S(\theta^\star)$, and its flow is $\Flow_t(s)=s^\star+e^{-t}(s-s^\star)$. For $s\in L$
and $t\ge0$, invariance gives $s_t\in L$ with
$\Flow_t(s_t)=s$, and
\begin{align*}
  \big\|s-s^\star\big\|
  \;=\;\big\|\Flow_t(s_t)-s^\star\big\|
  \;=\;e^{-t}\,\big\|s_t-s^\star\big\|
  \;\le\;e^{-t}\,\mathrm{diam}(\mathcal S)
  \;\xrightarrow[t\to\infty]{}\;0 .
\end{align*}
Thus $L=\{s^\star\}$, i.e., $s^{(k)}\to s^\star$.
\end{proof}

\subsubsection*{Proof of Theorem~\ref{theo:inexact} (Persistent inner-loop inaccuracy)}

\begin{proof}
The assertions are proven with $C=\cgrad\Lip{S}/\cdec$, where $\cdec,\cgrad$ are the constants of
\eqref{eq:F_bounds} and $\Lip{S}$ that of $\|\Delta^{\mathrm{in}}_k\|\le\Lip{S}\delta_k$.
Write $G_k=\|\nabla_\theta\mathcal L^\star(\theta^{(k)})\|$ and
$\xi_k=\langle\nabla\Lyap(s^{(k-1)}),\Delta^{\mathrm{mc}}_k\rangle$. The per-step inequality
\eqref{eq:per_step} with the sharper bound
$\|\nabla\Lyap(s^{(k-1)})\|\le\cgrad G_{k-1}$ from \eqref{eq:F_bounds} gives
\begin{align}
  \Lyap\big(s^{(k)}\big)
  \;\le\;\Lyap\big(s^{(k-1)}\big)-\gamma_k\,\cdec\,G_{k-1}^2
  +\gamma_k\,\cgrad\,G_{k-1}\Lip{S}\delta_k+\gamma_k\xi_k+c_2\gamma_k^2 .
  \label{eq:descent_bias}
\end{align}

\emph{Floor \eqref{eq:inexact_floor}.} Suppose it fails on an event of positive probability. Then
there are $\epsilon_1>0$, $\eta>\cgrad(\Lip{S}\delta+\epsilon_1)/\cdec$ and $k_0$ with
$G_k\ge\eta$ and $\Lip{S}\delta_k\le\Lip{S}\delta+\epsilon_1$ for all $k\ge k_0$. Hence, for
$k>k_0$,
\begin{align*}
  -\cdec\,G_{k-1}^2+\cgrad\,G_{k-1}\Lip{S}\delta_k
  &\;\le\;-G_{k-1}\Big(\cdec\,G_{k-1}-\cgrad\big(\Lip{S}\delta+\epsilon_1\big)\Big)\\
  &\;\le\;-\eta\Big(\cdec\,\eta-\cgrad\big(\Lip{S}\delta+\epsilon_1\big)\Big)
  \;=:\;-\epsilon_0\;<\;0 ,
\end{align*}
and summing \eqref{eq:descent_bias} gives
\begin{align*}
  \Lyap\big(s^{(n)}\big)
  \;\le\;\Lyap\big(s^{(k_0)}\big)-\epsilon_0\sum_{k=k_0+1}^{n}\gamma_k
  +\mathcal Z_n-\mathcal Z_{k_0}+c_2\sum_{k>k_0}\gamma_k^2
  \;\xrightarrow[n\to\infty]{}\;-\infty ,
\end{align*}
since the martingale $\mathcal Z_n=\sum_{k\le n}\gamma_k\xi_k$ converges, $\sum_k\gamma_k^2<\infty$
and $\sum_k\gamma_k=\infty$. This contradicts the boundedness of $\Lyap$ on $\mathcal S$.

\emph{Neighborhood \eqref{eq:inexact_ball}.} Since $\Lyap(s^{(k)})=-\mathcal L^\star(\theta^{(k)})$,
adding $\mathcal L^\star(\theta^\star)$ to \eqref{eq:descent_bias} gives \eqref{eq:capture_descent}
for $e_k=\mathcal L^\star(\theta^\star)-\mathcal L^\star(\theta^{(k)})$ with
\begin{align*}
  b_k\;=\;\Lip{S}\delta_k,
  \qquad
  \limsup_{k\to\infty}b_k\;\le\;\Lip{S}\delta\;=\;b,
  \qquad
  C_b\;=\;\frac{2\,\cgrad\,\Lip{S}}{\cdec\,m}\,\delta\;=\;\frac{2C}{m}\,\delta\;=\;C_\delta ,
\end{align*}
and $\sum_k\gamma_k\xi_k=\lim_n\mathcal Z_n$ converges. Since
$\|\theta^{(k)}-\theta^{(k-1)}\|\le\Lip{\mathcal M}c_g\gamma_k\to0$ as in the proof of
Theorem~\ref{theo:convergence}, Lemma~\ref{lemma:capture} applies with $f=\mathcal L^\star$ and
$\Theta'=\Theta_{\mathcal S}$.
\end{proof}

\noindent
For Theorem~\ref{theo:is}, Lemma~\ref{lemma:is_consistent} is proven first, and the properties of
the importance-weighted recursion are collected.

\subsubsection*{Proof of Lemma~\ref{lemma:is_consistent} (Consistency of the weighted statistics)}
\begin{proof}
Write $w_i(z)=p(x_i,z;\theta)/r_i(z)$ for the unnormalized weights and $u$ for a component of
\eqref{eq:complete_statistics}, i.e., of $z_i$, $z_iz_i^\top$ or $\sum_lr_{il}(z_i)^2$. By
Assumption~\ref{ass:regularity}(i), $p(x_i|z;\theta)$ is bounded, so $p(x_i,z;\theta)$ has the
Gaussian tail of the prior, whereas the $t$ density $r_i$ decays only polynomially. Hence $w_i$ and
$w_iu$ are bounded on $\Reals^{n_z}$. Moreover, $|u(z)|\le c(1+\|z\|^2)$, so $u$ has finite moments
of order $2p$ under $r_i$ for every $p<\nu/4$, in particular for some $p>1$, since $\nu>4$. The strong
law of large numbers in numerator and denominator gives
\cite[Theorem~1]{GewekeBayesianInferenceEconometric89}
\begin{align*}
  \sum_{j=1}^K\tilde w_{ij}\,u(z_{ij})
  &\;=\;\frac{\frac1K\sum_{j=1}^K w_i(z_{ij})\,u(z_{ij})}{\frac1K\sum_{j=1}^K w_i(z_{ij})}
  \;\xrightarrow[K\to\infty]{\text{a.s.}}\;
  \frac{\Exp_{z\sim r_i}\big[w_i(z)\,u(z)\big]}{\Exp_{z\sim r_i}\big[w_i(z)\big]}\\
  &\;=\;\frac{\int u(z)\,p(x_i,z;\theta)\,dz}{\int p(x_i,z;\theta)\,dz}
  \;=\;\Exp_{z\sim p(\cdot|x_i;\theta)}\big[u(z)\big],
\end{align*}
and summing over $i$ gives $\widehat S_K\to\bar S_p(\theta)$ almost surely. For the first two
blocks, the bounds \eqref{eq:is_finite_K} follow from
\cite[Theorem~2.3]{AgapiouImportanceSamplingIntrinsic17}, which bounds the bias and the mean squared
error of self-normalized importance sampling by $1/K$ times a constant for unbounded test
functions. This constant is given by the moments of $u$ of order $2p$ and of $w_iu$ and $w_i$
under $r_i$, which are finite by the above. For the bounded third block,
\cite[Theorem~2.1]{AgapiouImportanceSamplingIntrinsic17} gives a constant proportional to
$\rho(\theta,\vq)$. All moments depend continuously on $(\theta,\vq)$ by dominated convergence,
so $c_{\mathrm{is}}$ is bounded on compact subsets of $\Theta\times\mathcal Q$.
\end{proof}

\begin{lemma}[Importance-weighted recursion]
\label{lemma:is_rec}
Let the assumptions of Theorem~\ref{theo:is} hold. Then
\begin{enumerate}
\item[(i)] $\Delta^{\mathrm{mc}}_k$ is a martingale difference, $\Exp[\Delta^{\mathrm{mc}}_k\mid\mathcal
      G_k]=0$, with $\Exp[\|\Delta^{\mathrm{mc}}_k\|^2\mid\mathcal G_k]\le\bar c_{\mathrm{is}}/K$, and the
      bias satisfies
      \begin{align}
        \label{eq:is_bias}
        \big\|\Delta^{\mathrm{is}}_k\big\|\;\le\;\frac{\bar c_{\mathrm{is}}}{K},
        \qquad
        \bar c_{\mathrm{is}}\;=\;\sup\big\{c_{\mathrm{is}}(\theta,\vq):\ \theta\in\Theta_{\mathcal S_p},\
        \vq\in\mathcal Q\big\}\;<\;\infty ,
      \end{align}
      with $c_{\mathrm{is}}$ from Lemma~\ref{lemma:is_consistent},
\item[(ii)] $\Lyap_p$ is continuously differentiable with Lipschitz gradient on $\mathcal S_p$,
      \begin{align}
        \label{eq:decay_p}
        \big\langle\nabla_s\Lyap_p(s),\,\Fmf_p(s)\big\rangle
        \;=\;-\,\big\|\nabla_\theta\mathcal{LL}(\theta_s)\big\|^2_{H_s^{-1}}\;\le\;0 ,
      \end{align}
      and the bounds \eqref{eq:F_bounds} hold with $\mathcal{LL}$ in place of $\mathcal L^\star$, with
      constants $\cdec,\cgrad$ taken over $\mathcal S_p$.
\end{enumerate}
\end{lemma}

\begin{proof}
\emph{(i).} Conditionally on $\mathcal G_k$, the iterate $\theta^{(k-1)}\in\Theta_{\mathcal S_p}$ and
the encoder output $\vq^{(k)}\in\mathcal Q$ are fixed. The conditional variance is bounded by the
mean squared error, so both bounds are \eqref{eq:is_finite_K} conditionally on $\mathcal G_k$. The
supremum is finite, since $\Theta_{\mathcal S_p}=\mathcal M(\mathcal S_p)$ is compact as the
continuous image of a compact set (Assumptions~\ref{ass:mstep} and~\ref{ass:is}) and $\mathcal Q$ is compact.
\emph{(ii).} This is the computation of Lemma~\ref{lemma:lyapunov} with the Fisher identity
$\nabla_\theta\mathcal{LL}(\theta)=\Dzeta(\theta)^\top\bar S_p(\theta)-\nabla_\theta\kappa(\theta)$ in
place of the envelope identity. By $\nabla_s\mathcal M(s)=H_s^{-1}\Dzeta(\theta_s)^\top$ and the chain
rule, $\nabla_s\Lyap_p(s)=-\Dzeta(\theta_s)H_s^{-1}\nabla_\theta\mathcal{LL}(\theta_s)$, and
\begin{align*}
  \Dzeta(\theta_s)^\top\big(\bar S_p(\theta_s)-s\big)
  &\;=\;\big[\Dzeta(\theta_s)^\top\bar S_p(\theta_s)-\nabla_\theta\kappa(\theta_s)\big]
     -\big[\Dzeta(\theta_s)^\top s-\nabla_\theta\kappa(\theta_s)\big]\\
  &\;=\;\nabla_\theta\mathcal{LL}(\theta_s)-0 ,
\end{align*}
where the second bracket vanishes by the first-order condition \eqref{eq:mstep_foc} of the M-step. Taking the inner product with
$-H_s^{-1}\nabla_\theta\mathcal{LL}(\theta_s)$ gives \eqref{eq:decay_p}. The constants follow from
Lemma~\ref{lemma:mop}(i) as in Lemma~\ref{lemma:lyapunov}.
\end{proof}

\subsubsection*{Proof of Theorem~\ref{theo:is} (Convergence of the VAE-NLME algorithm)}
\begin{proof}
The assertions are proven with $C=\cgrad\bar c_{\mathrm{is}}/\cdec$, where $\cdec,\cgrad$ are the
constants of Lemma~\ref{lemma:is_rec}(ii) and $\bar c_{\mathrm{is}}$ that of
Lemma~\ref{lemma:is_rec}(i). Write $g_k=\Fmf_p(s^{(k-1)})+\Delta^{\mathrm{mc}}_k+\Delta^{\mathrm{is}}_k$,
so that $s^{(k)}=s^{(k-1)}+\gamma_kg_k$. In contrast to the variational statistics, $g_k$ is not
bounded, but by Lemma~\ref{lemma:is_rec}(i) its conditional second moment is, since $\Fmf_p$ is
bounded on the compact set $\mathcal S_p$. With Assumption~\ref{ass:stepsizes},
\begin{align}
  \label{eq:is_squares}
  \Exp\Big[\sum_{k\ge1}\gamma_k^2\|g_k\|^2\Big]\;<\;\infty,
  \qquad\text{hence}\qquad
  \sum_{k\ge1}\gamma_k^2\|g_k\|^2\;<\;\infty
  \quad\text{and}\quad
  \gamma_k\|g_k\|\;\to\;0
\end{align}
almost surely. In particular, the steps $\|\theta^{(k)}-\theta^{(k-1)}\|\le\Lip{\mathcal
M}\gamma_k\|g_k\|$ vanish, with the Lipschitz constant of $\mathcal M$ on $\mathcal S_p$
(Lemma~\ref{lemma:mop}(ii) for $\mathcal S_p$). As in the proof of
Proposition~\ref{prop:discrete_decay}, the descent lemma and Lemma~\ref{lemma:is_rec} give
\begin{align}
  \label{eq:descent_is}
  \Lyap_p\big(s^{(k)}\big)
  \;\le\;\Lyap_p\big(s^{(k-1)}\big)-\gamma_k\,\cdec\,G_{k-1}^2
  +\gamma_k\,\cgrad\,G_{k-1}\big\|\Delta^{\mathrm{is}}_k\big\|+\gamma_k\xi_k
\end{align}
with $G_k=\|\nabla_\theta\mathcal{LL}(\theta^{(k)})\|$ and
$\xi_k=\langle\nabla\Lyap_p(s^{(k-1)}),\Delta^{\mathrm{mc}}_k\rangle
+\tfrac{\Lip{\nabla\Lyap_p}}{2}\gamma_k\|g_k\|^2$. The series $\sum_k\gamma_k\xi_k$ converges almost
surely, its first part as a martingale with square-summable increments, its second part by
\eqref{eq:is_squares}.

\emph{Assertion (i).} Since $\|\Delta^{\mathrm{is}}_k\|\le\bar c_{\mathrm{is}}/K$ by
Lemma~\ref{lemma:is_rec}(i), \eqref{eq:descent_is} is \eqref{eq:descent_bias} with $\Lip{S}\delta_k$
replaced by $\bar c_{\mathrm{is}}/K$ and $c_2=0$. Hence, the proof of the floor
\eqref{eq:inexact_floor} gives $\liminf_kG_k\le\cgrad\bar c_{\mathrm{is}}/(\cdec K)=C/K$.

\emph{Assertion (ii).} For $K_k\to\infty$ with $\sum_k\gamma_k/K_k<\infty$, the bias bound gives
\begin{align*}
  \big\|\Delta^{\mathrm{is}}_k\big\|\;\le\;\frac{\bar c_{\mathrm{is}}}{K_k}\;\xrightarrow[k\to\infty]{}\;0,
  \qquad
  \sum_k\gamma_k\big\|\Delta^{\mathrm{is}}_k\big\|\;\le\;\bar c_{\mathrm{is}}\,\sum_k\frac{\gamma_k}{K_k}\;<\;\infty .
\end{align*}
Since $G_k$ is bounded on the compact set $\Theta_{\mathcal S_p}$, summing \eqref{eq:descent_is}
gives $\sum_k\gamma_kG_{k-1}^2<\infty$, as in Proposition~\ref{prop:discrete_decay}. The proof of
Proposition~\ref{prop:F_to_zero} then carries over, with \eqref{eq:F_increments} replaced by
\begin{align*}
  \big|R_m-R_n\big|
  \;\le\;\Lip{R}\big\|s^{(m)}-s^{(n)}\big\|
  \;\le\;\Lip{R}\Big(c\sum_{k=n+1}^{m}\gamma_k
  +\sup_{l>n}\Big\|\sum_{k=n+1}^{l}\gamma_k\Delta^{\mathrm{mc}}_k\Big\|\Big)
\end{align*}
with $c=\sup_{\mathcal S_p}\|\Fmf_p\|+\bar c_{\mathrm{is}}$, whose last term tends to zero almost surely as $n\to\infty$, since $\sum_k\gamma_k\Delta^{\mathrm{mc}}_k$
converges. Hence $G_k\to0$. Since $\nabla_\theta\mathcal{LL}$ is continuous and $\Theta_{\mathcal S_p}$
is compact, $\theta^{(k)}$ converges to the set of stationary points of $\mathcal{LL}$ in
$\Theta_{\mathcal S_p}$, and to $\theta_{\mathrm{ML}}$ if it is the only one.

\emph{Assertion (iii).} Adding $\mathcal{LL}(\theta_{\mathrm{ML}})$ to \eqref{eq:descent_is} gives
\eqref{eq:capture_descent} for $e_k=\mathcal{LL}(\theta_{\mathrm{ML}})-\mathcal{LL}(\theta^{(k)})$
with $b_k=\|\Delta^{\mathrm{is}}_k\|$, $c_2=0$ and the noise $\xi_k$ from above. The steps vanish by
\eqref{eq:is_squares}, and $\mathcal{LL}$ is twice continuously differentiable on $\Theta$ by
differentiation under the integral, since the $\theta$-derivatives of $p(x,z;\theta)$ are dominated
by a polynomial in $z$ times the Gaussian tail of the prior, locally uniformly in $\theta$. Hence
Lemma~\ref{lemma:capture} applies with $f=\mathcal{LL}$ and $\Theta'=\Theta_{\mathcal S_p}$, and the
bias bound gives
\begin{align*}
  b\;=\;\frac{\bar c_{\mathrm{is}}}{K},\quad C_b\;=\;C_K
  \qquad\text{for fixed }K,
  \qquad\qquad
  b\;=\;0,\quad C_b\;=\;0
  \qquad\text{for }K_k\to\infty .
\end{align*}
\end{proof}

\subsection{Residual Evaluation}
\label{app:residuals}

The residuals are computed at every iterate. The residual $\|\nabla_\theta\mathcal L^\star\|$ of
Theorem~\ref{theo:convergence} follows from the envelope identity \eqref{eq:envelope}, with
$\vq^\star(\theta)$ obtained by free optimization over $(\mu_i,L_i)$, and the residual
$\|\nabla_\theta\mathcal{LL}\|$ of Theorem~\ref{theo:is} from the Fisher identity \eqref{eq:fisher}.
Both require expectations over the individual parameters. For a Gaussian
$\mathcal N(\mu_i,L_iL_i^\top)$ and a function $g$, they are evaluated by Gauss-Hermite quadrature
instead of a Monte Carlo sum,
\begin{align*}
  \text{Gauss-Hermite:}&\quad
  \Exp_{\varepsilon\sim\mathcal N(0,I)}\big[g(\mu_i+L_i\varepsilon)\big]
  \;\approx\;\sum_{j\in\{1,\dots,n\}^{n_z}}\omega_j\,g\big(\mu_i+L_iu_j\big),\\
  \text{Monte Carlo:}&\quad
  \Exp_{\varepsilon\sim\mathcal N(0,I)}\big[g(\mu_i+L_i\varepsilon)\big]
  \;\approx\;\frac1M\sum_{m=1}^M g\big(\mu_i+L_i\varepsilon^{(m)}\big),
  \qquad \varepsilon^{(m)}\sim\mathcal N(0,I),
\end{align*}
where $u_j=(u_{j_1},\dots,u_{j_{n_z}})$ and $\omega_j=\omega_{j_1}\cdots\omega_{j_{n_z}}$ are the
nodes and weights of the product rule built from the $n$-point Gauss-Hermite rule for
$\mathcal N(0,1)$. Expectations under the posterior distribution use the same grid on the widened
encoder distribution $\mathcal N(\mu_i,1.6^2L_iL_i^\top)$ with self-normalized weights as in
\eqref{eq:is_statistics}.
\\
Since $z_i$ is only three-dimensional, the grid has $n^3$ points, i.e., $1728$ for the $n=12$
nodes of the ELBO and between $21952$ and $46656$ for the $n=28$ to $n=36$ nodes of the
log-likelihood. The Monte Carlo
error decays only like $M^{-1/2}$
\cite{KleywegtSampleAverageApproximation02,BurroniDeterministicObjectiveBBVI24}, whereas $12$ nodes
per dimension agree to ten digits with a Monte Carlo average over $2^{20}$ samples. Hence, in this
setting, Gauss-Hermite quadrature is more accurate at a lower cost. 

\end{document}

%% file: graphics/numbers.tex
\newcommand{\NumFloorLL}{4.8\cdot 10^{-2}}
\newcommand{\NumFloorLstar}{4.8\cdot 10^{-2}}
\newcommand{\NumGvar}{0.06}

%% file: graphics/story_numbers.tex
\newcommand{\NumStPracLL}{335.7}

\newcommand{\NumStPracN}{205}

\newcommand{\NumStElboLL}{335.9}

%% file: graphics/theory_numbers.tex
\newcommand{\NumThLyapVAfterKL}{6.5\cdot 10^{-7}}

%% file: graphics/encoder_numbers.tex
\newcommand{\NumEncStartDist}{0.13}

\newcommand{\NumEncPlateauEnc}{70}
\newcommand{\NumEncPlateauFree}{160}
\newcommand{\NumEncTab}{%
ELBO VAE-NLME, encoder & $-171.19$ & $-168.02$ & $-167.97$ & $3.16$ & $0.06$ & $3.5$ & $3.5$ \\
ELBO VAE-NLME, free $\vq$ & $-170.42$ & $-167.93$ & $-167.88$ & $2.49$ & $0.06$ & $2.1$ & $2.1$ \\
VAE-NLME, encoder & $-169.94$ & $-167.89$ & $-167.83$ & $2.05$ & $0.06$ & $2.4\cdot 10^{-1}$ & $2.5\cdot 10^{-1}$ \\
\hline
$\theta^\star_{\mathrm{ELBO}}$ & $\cdot$ & $-167.89$ & $-167.83$ & $\cdot$ & $0.06$ & $4.9\cdot 10^{-6}$ & $4.8\cdot 10^{-2}$ \\
$\theta_{\mathrm{ML}}$ & $\cdot$ & $-167.89$ & $-167.83$ & $\cdot$ & $0.06$ & $4.8\cdot 10^{-2}$ & $5.9\cdot 10^{-7}$ \\}